\documentclass[sigconf]{acmart}

\AtBeginDocument{%
  }
\usepackage{multirow}
\usepackage{bbding}
\setcopyright{acmlicensed}
\copyrightyear{2024}
\acmYear{2024}
\acmDOI{10.1145/3664647.3681022}

\acmISBN{979-8-4007-0686-8/24/10}

\begin{document}

\title{3D Question Answering for City Scene Understanding}


\author{Penglei Sun}
\orcid{0009-0005-2290-0944}
\affiliation{%
  \institution{The Hong Kong University of Science and Technology (Guangzhou)}
  \city{Guangzhou}
  \country{China}
  }
\email{psun012@connect.hkust-gz.edu.cn}
\authornote{Equal Contribution.}

\author{Yaoxian Song}
\orcid{0000-0002-8146-2236}
\affiliation{%
  \institution{Zhejiang University}
  \city{Hangzhou}
  \country{China}
  }
\email{songyaoxian@zju.edu.cn}
\authornotemark[1]

\author{Xiang Liu}
\orcid{0009-0003-2639-1804}
\affiliation{%
  \institution{The Hong Kong University of Science and Technology (Guangzhou)}
  \city{Guangzhou}
  \country{China}
  }
\email{xliu886@connect.hkust-gz.edu.cn}

\author{Xiaofei Yang}
\orcid{0000-0003-2458-6774}
\affiliation{%
  \institution{Guangzhou University}
  \city{Guangzhou}
  \country{China}
  }
\email{xiaofeiyang@gzhu.edu.cn}

\author{Qiang Wang}
\orcid{0000-0002-2986-967X}
\affiliation{%
  \institution{Harbin Institute of Technology (Shenzhen)}
  \city{Shenzhen}
  \country{China}
  }
\email{qiang.wang@hit.edu.cn}
\authornote{Corresponding author.}

\author{Tiefeng Li}
\orcid{0000-0003-0265-3454}
\affiliation{%
  \institution{Zhejiang University}
  \city{Hangzhou}
  \country{China}
  }
\email{litiefeng@zju.edu.cn}

\author{Yang Yang}
\orcid{0000-0003-0608-9408}
\affiliation{%
  \institution{The Hong Kong University of Science and Technology (Guangzhou)}
  \city{Guangzhou}
  \country{China}
  }
\email{yyiot@hkust-gz.edu.cn}

\author{Xiaowen Chu}
\orcid{0000-0001-9745-4372}
\affiliation{%
  \institution{The Hong Kong University of Science and Technology (Guangzhou)}
  \city{Guangzhou}
  \country{China}
}
\email{xwchu@ust.hk}
\authornotemark[2]
\newcommand{\syx}[1]{{\color{black} #1}}
\renewcommand{\shortauthors}{Penglei Sun et al.}

\begin{abstract}
\syx{
3D multimodal question answering (MQA) plays a crucial role in scene understanding by enabling intelligent agents to comprehend their surroundings in 3D environments.
While existing research has primarily focused on indoor household tasks and outdoor roadside autonomous driving tasks, there has been limited exploration of city-level scene understanding tasks. 
Furthermore, existing research faces challenges in understanding city scenes, due to the absence of spatial semantic information and human-environment interaction information at the city level.
To address these challenges, we investigate 3D MQA from both dataset and method perspectives.
From the dataset perspective, we introduce a novel 3D MQA dataset named \underline{\textbf{City-3DQA}} for city-level scene understanding, 
which is the first dataset to incorporate scene semantic and human-environment interactive tasks within the city.
From the method perspective, we propose a \textbf{S}cene \textbf{g}raph enhanced \textbf{City}-level \textbf{U}nderstanding method (\underline{\textbf{Sg-CityU}}), which utilizes the scene graph to introduce the spatial semantic. 
A new benchmark is reported and our proposed Sg-CityU achieves accuracy of $63.94 \%$ and $63.76 \%$ in different settings of City-3DQA.
Compared to indoor 3D MQA methods and zero-shot using advanced large language models (LLMs), Sg-CityU demonstrates state-of-the-art (SOTA) performance in robustness and generalization. 
Our dataset and code are available on our project website\footnote{\url{https://sites.google.com/view/city3dqa/}}.
}


\end{abstract}

\ccsdesc[500]{Computing methodologies~Scene understanding}

\keywords{multimodal question answering, scene understanding, 3D}


\maketitle

\section{Introduction}

\begin{figure}[t]
  \centering
  \includegraphics[width=0.8\linewidth]{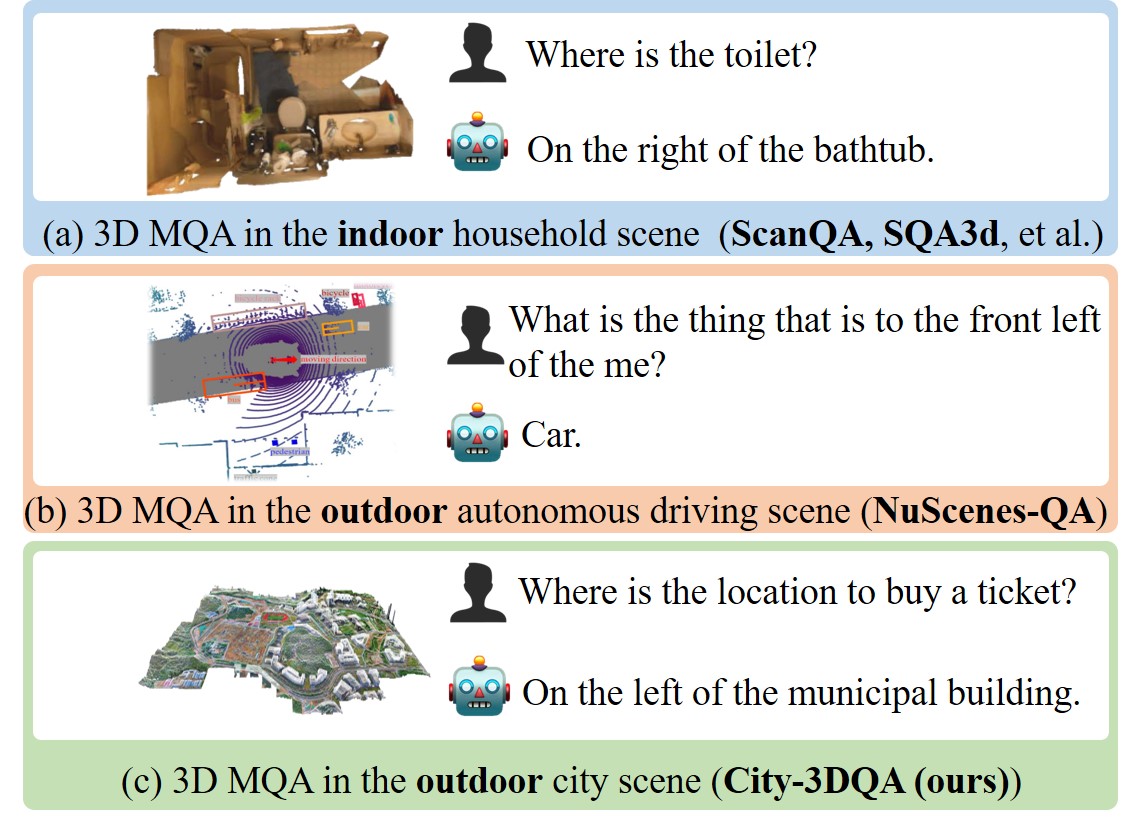}
  \caption{Comparison of the City-3DQA with other 3D multimodal question answering (MQA) tasks.
  The existing research in 3D MQA focuses on the indoor household scene (a) and outdoor autonomous driving scene (b).
  However, these researches lack spatial semantic and city-level interaction information within the city.
  City-3DQA (c) is the first dataset to focus on 3D MQA for outdoor city scene understanding.}
  \label{fig:pipeline_overview}
\end{figure}

City scene understanding is a crucial technology for guided tour~\cite{wallgrun2020comparison}, autonomous systems~\cite{Goddard2021}, and smart city~\cite{chan2016tackling}. 
\syx{3D multimodal question answering (MQA) is one of the key manners of human-environment interaction to promote city scene understanding~\cite{lee2021towards}.}
\syx{For instance, people with visual impairment could interact with the electronic personal assistant (seen as an agent) integrated into wearable smart glasses, such as Microsoft HoloLens~\cite{Microsof28:online} or Apple Vision Pro~\cite{AppleVis63:online}, to obtain auxiliary scenario information in the situated city by asking questions with city perception from the embedded visual sensors, shown in Figure~\ref{fig:pipeline_overview} (c). }


\begin{figure*}[t]
  \centering
  \includegraphics[width=0.8\linewidth]{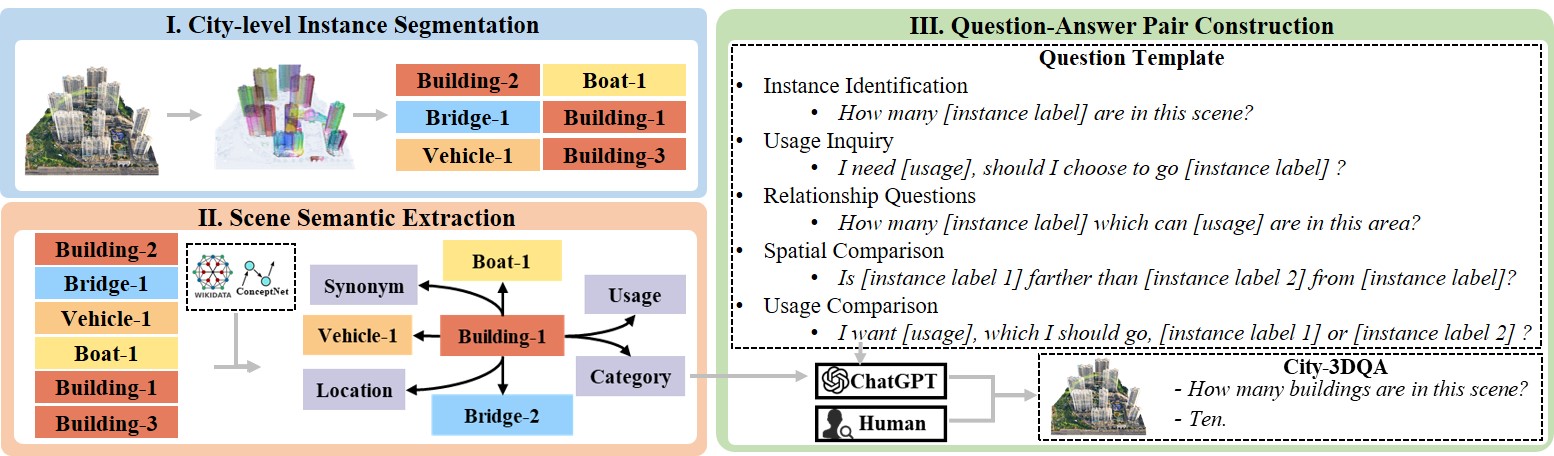}
  \caption{Data Construction Pipeline for City-3DQA. The pipeline consists of three main stages: City-level Instance Segmentation, Scene Semantic Extraction, and Question-Answer Pair Construction. 
  }
\label{fig:dataset_pipeline}
\end{figure*}

However, existing 3D MQA tasks face challenges in city scene understanding due to lacking spatial semantic information and city-level interaction information within the city, such as the location and the usage of instances.
Existing research mainly focuses on two lines including the 3D MQAs in the indoor household setting (Fig.~\ref{fig:pipeline_overview} (a)) and the 3D MQAs in the outdoor autonomous driving settings(Fig.~\ref{fig:pipeline_overview} (b)). 
For the former, EQA~\cite{das2018embodied}, MP3D-EQA~\cite{wijmans2019embodied}, MT-EQA~\cite{yu2019multi} and EMQA~\cite{datta2022episodic} realize MQA-based scene understanding using images in indoor household scenarios through House3D simulation environment~\cite{wu2018building} for navigation tasks.
Apart from using images, there is also 3D MQA research, such as 3DQA~\cite{ye20223d}, ScanQA~\cite{azuma2022scanqa}, CLEvR3D~\cite{yan2023comprehensive}, FE-3DGQA~\cite{zhao2022towards} and SQA3D~\cite{ma2022sqa3d}, which adopt point cloud for indoor household scene understanding based on the point cloud environment ScanNet~\cite{dai2017scannet}.
For the latter, \citet{qian2024nuscenes} introduce NuScenes-QA in outdoor settings firstly for autonomous driving using the point cloud.
This task focuses on roadside-related instances including cars and pedestrians, yet it does not consider other instances in the city such as \textit{plantings}, \textit{buildings}, and \textit{rivers}.
In summary, current 3D MQAs are hard to satisfy city-level scene understanding for urban activities of humans or agents.

To address these challenges, we explore the task from both the dataset and method perspectives.
From the dataset perspective, we introduce \textbf{City-3DQA}, the first 3D MQA dataset for outdoor city scene understanding in Figure~\ref{fig:dataset_pipeline}.
We realize data collection including City-level Instance Segmentation, Scene Semantic Extraction, and Question-Answer Pair Construction.
Specifically, in City-level Instance Segmentation, we utilize pre-trained instance segmentation models to identify city instances. 
In Scene Semantic Extraction, we construct the scene semantic information for instances in the graph structure, including spatial information and semantic information. 
The spatial information denotes relationships between pairs of instances, such as "\textit{living building - left - business building}".
The semantic information represents instances with attributes, such as "\textit{transportation building - usage - buying tickets}".
In Question-Answer Pair Construction, we develop $33$ unique question templates that enable multi-hop reasoning and urban activities, which are classified into five categories: instance identification, usage inquiry, relationship questions, spatial comparison, and usage comparison for the city scene understanding inspired by ~\citet{gao2022cric} and ~\citet{qian2024nuscenes}.
The LLM leverages these templates in combination with scene semantic information to produce question-answer pairs.
The human evaluation assesses dataset quality. 
The City-3DQA dataset comprises $\mathbf{450}$\textbf{k question-answer pairs} and $\mathbf{2.5}$ \textbf{billion point clouds} across six cities.

From the method perspective, we introduce a \textbf{S}cene \textbf{g}raph enhanced \textbf{City}-level \textbf{U}nderstanding method (\textbf{Sg-CityU}) for City-3DQA.
Compared to indoor scene understanding, city-level scene understanding is limited by sparse semantic information due to large scales.
This leads to challenges associated with long-range connections and spatial inference during the modeling process~\cite{liao2022kitti}.
Therefore, Sg-CityU utilizes the scene graph to introduce spatial relationship information among instances.
Specifically, for the input point cloud and the question, Sg-CityU extracts the vision and language representation from point clouds and questions respectively.
And then a city-level scene graph is constructed, which is encoded through graph neural networks~\cite{kipf2016semi,kim20193}.
We design the Fusion Layer to fuse aforementioned scene multimodal representations for answering generation.

Our main contributions can be summarized as follows:

\begin{enumerate}

\item We investigate 3D multimodal question answering (MQA) to realize city-level scene understanding for urban activities of humans or agents.

\item We introduce a novel large-scale dataset named City-3DQA. 
To our knowledge, City-3DQA is the first dataset to consider scene semantic information and city-level interactive tasks.

\item We provide a baseline method (\textbf{Sg-CityU}), which introduces spatial relationship information through the scene graph to generate high-quality city-related answers.   

 
\item A new benchmark is proposed in which evaluations are conducted with existing MQA methods and LLM-based zero-shot methods on our City-3DQA. Experimental results show that our proposed \textbf{Sg-CityU} achieves the best performance in robustness and generalization, specifically, $63.94 \%$ and $63.76 \%$ accuracy in sentence-wise and city-wise settings respectively.



\end{enumerate}

\section{Related Work}

\subsection{City Scene Understanding}
Existing research in city scene understanding primarily concentrates on segmentation, reconstruction, and grounding.
City segmentation, as explored in works such as~\citet{liao2022kitti,yang2023urbanbis,hu2022sensaturban,geng20233dgraphseg}, aims to distinguish different instances within city-level point clouds or meshes for a comprehensive understanding of urban environments.
City scene reconstruction, as discussed in~\citet{tang2022point,zhang2021continuous,lin2022capturing,kuang2020real}, seeks to understand the visual information of each object in city scenes and reconstruct their geometries from partial observations, such as point clouds from 3D scans.
However, these methods primarily focus on visual representation rather than language representation and semantic information in city scenes, which are important for human-environment interaction. 
~\citet{miyanishi2023cityrefer} introduce CityRefer, which addresses city-level visual grounding by localizing objects in 3D scenes based on language expressions.
Inspired by these studies, our research aims to tackle this problem from a multimodal question answering perspective. 
We propose the first 3D multimodal question answering dataset, City-3DQA, for 3D city scene understanding, which integrates language representation and semantic information.

\subsection{3D Multimodal Question Answering}
3D Multimodal Question Answering is a novel task within the field of scene understanding, concentrating on the ability to answer questions about 3D scenes, which are depicted through simulated environments or point clouds~\cite{azuma2022scanqa}.
~\citet{das2018embodied,wijmans2019embodied,datta2022episodic,yu2019multi} present an embodied question answering where the agent must first intelligently navigate to explore the environment, gather the necessary visual information through first-person vision, and then respond to the question in a 3D simulated environment.
~\citet{ye20223d,etesam20223dvqa,azuma2022scanqa,yan2023comprehensive,zhao2022towards,ma2022sqa3d} propose a series of studies based on the ScanNet dataset~\cite{dai2017scannet} that focus on processing point cloud data from entire 3D indoor scenes to respond to specific textual queries about the environment.
However, these works focus on the indoor household scene and overlook the outdoor scene.
~\citet{qian2024nuscenes} proposes the outdoor 3D multimodal question NuScenes-QA answering benchmark to address the human-machine interaction in autonomous driving rather than the city scene understanding.
We first introduce City-3DQA, a 3D question-answering dataset specifically designed for the understanding of outdoor city scenes. 
Unlike the NuScenes-QA which concentrates on roadside areas, City-3DQA emphasizes the comprehension of city landscapes along with their spatial characteristics. 
Additionally, it incorporates features related to interaction, such as usage.

\section{Problem Definition}
The 3D MQA for city scene understanding is formulated as follows: given inputs of the point cloud $p$ and question $q$ about the 3D city scene, the model aims to output $\widehat{a}$ that semantically matches true answer $a^{*}$ from the answer set $\mathbb{A}$,
\begin{equation}
\small
\scalebox{0.95}{$
\begin{aligned} 
\widehat{a} = \mathop{\arg \max }\limits_{a \in \mathbb{A}} \; \text{P}(a | p , q).
\end{aligned}
$}
\label{equ:task_def}
\end{equation}

Understanding city-level scenes is more challenging than indoor scenes. 
This is because city scenes have less dense information over large areas, making it hard to model long-range connections and spatial relationships~\cite {liao2022kitti}.
Therefore, we introduce a scene graph $sg$ which contains the relative spatial relationship~\cite{xu2020survey}.
The $sg$ is composed of nodes and edges, where the nodes represent instances and the edges represent the spatial relationships between these instances.
We consider a scene-graph-aware joint probability model for the task using $sg$ and decompose Equation~\ref{equ:task_def} into two parts,, given by:
\begin{equation}
\small
\scalebox{1.0}{$
\begin{aligned} 
\text{P}(a | p , q) = \text{P}(a | p , q, sg) \times \text{P}(sg | p).
\end{aligned}
$}
\label{equ:main_equation}
\end{equation}

\section{City-3DQA Dataset}

\begin{table*}[t]
  \caption{Comparison between City-3DQA and other 3D MQA datasets. 
  Question-Answer Pairs and Point Clouds denote the number of question-answer pairs and points.}
  \label{tab:comparison_vqa}
  \scalebox{0.70}{
  \begin{tabular}{@{}c|cccccc@{}}
\toprule
Dataset                                                   & Scene   & Collection              & Scale     & Input Modal     & Question-Answer Pairs & Point Clouds \\ \midrule
EQA~\cite{das2018embodied}          & indoor  & template                & Room     & image               & 1.5k     & -        \\
MP3D-EQA~\cite{wijmans2019embodied} & indoor  & template                & Room     & image               & 1.1k     & -        \\
EMQA~\cite{datta2022episodic}       & indoor  & human                   & Room     & image               & 9.7k     & -        \\
MT-EQA~\cite{yu2019multi}           & indoor  & template+human          & Room     & image               & 19k      & -        \\
3DVQA~\cite{etesam20223dvqa}        & indoor  & template                & Room     & point cloud         & 484k     & 242M     \\
3DQA~\cite{ye20223d}                & indoor  & human                   & Room     & point cloud         & 10k      & 242M     \\
ScanQA~\cite{azuma2022scanqa}       & indoor  & auto + human            & Room     & point cloud         & 41k      & 242M     \\
CLEVR3D~\cite{yan2023comprehensive} & indoor  & template                & Room     & point cloud         & 60.1k    & 242M     \\
FE-3DGQA~\cite{zhao2022towards}     & indoor  & human                   & Room     & point cloud         & 20k      & 242M     \\
SQA3D~\cite{ma2022sqa3d}            & indoor  & human                   & Room     & point cloud + image & 33.4k    & 242M     \\ \midrule
NuScenes-QA~\cite{qian2024nuscenes} & outdoor & template                & Roadside & point cloud + image & 460k     & 1.4B     \\
\textbf{City-3DQA (ours)}                                   & outdoor & template + auto + human & City     & point cloud         & 450k     & 2.5B     \\ \bottomrule
\end{tabular}
}
\end{table*}

\subsection{Data Construction}
\label{section:data_construction}

We develop an automatic pipeline for the construction of the City-3DQA dataset, as depicted in Figure~\ref{fig:dataset_pipeline}. 
The City-3DQA dataset is derived from the 3D city point cloud dataset UrbanBIS~\cite{yang2023urbanbis}.
Our pipeline encompasses three primary components: City-level Instance Segmentation, Scene Semantic Extraction, and Question-Answer Pair Construction.

\textbf{City-level Instance Segmentation.} \quad
We use pre-trained instance segmentation~\cite{yang2023urbanbis} for the UrbanBIS dataset and obtain a wide range of city instances including buildings, vehicles, vegetation, roads, and bridges covering six cities, Qingdao, Wuhu,	Longhua, Yuehai, Lihu, and Yingrenshi.
We extract the instance-level label along with annotations and spatial locations to build the instance set $S_{I}=\{i, (x_i, y_i, z_i) | i \in I\}$ from UrbanBIS, where $I$ is the instances from the raw dataset.  
$x_i, y_i, z_i$ is the x-axis, y-axis, and z-axis coordinate for each $i$.

\textbf{Scene Semantic Extraction.} \quad
We construct the scene semantic information $G_{i}$ for each instance $i$ in the graph structure, which comprises two components: the spatial information $sp_{i}$ and the semantic information $se_{i}$ in the graph structure.
$sp_{i}$ contains a series triples $(i, r_{i,j}^{sp}, j)$, where $r_{i,j}^{sp}$ is the spatial relationship between the instances $(i, j)$, where $ i \in S_{I}, j \in S_{I}$. 
These relationships are centered around instance $i$ and we define counterclockwise as the positive direction.
$R_{i,j}$ are divided via eight relationships: “\textit{front}”, “\textit{front-right}”, “\textit{right}”, “\textit{back-right}”, "\textit{front-left}", "\textit{left}", "\textit{back-left}" and "\textit{back}", depending on relative instance spatial positions and the angle $\theta = \arctan \frac{y_j - y_i}{x_j - x_i}$ between instance $i$ and $j$,


$se_{i}$ are defined as triples $(i, r_{i}^{se}, v_{i})$, where $r_{i}^{se}$ and $v_{i}$ are the attribute and value for instance $i$ respectively.
In City-3DQA, we define $r_{i}^{se}$ as five attributes including instance label, building category label, synonym label, location, and usage label. 
The instance label and a detailed building category label are sourced from the pre-trained instance segmentation method~\cite{yang2023urbanbis}. 
Drawing inspiration from ~\citet{henderson2016building}, we acknowledge the usage label as an important aspect of urban activities within the city scene. 
To enhance the relevance of the City-3DQA datasets to a common language and to promote linguistic variety, we integrate synonyms, as suggested by~\cite{schotter2013synonyms}.
The sources for usage descriptions and synonym labels are knowledge base WikiData~\cite{vrandevcic2014wikidata} and ConceptNet~\cite{speer2017conceptnet}.

\textbf{Question-Answer Pair Construction.} \quad
To construct the question-answer pairs automatically, we propose a template-based pipeline utilizing LLM to transform structured data $G_{i}$ into unstructured language question $q_{i}$ and answer $a_{i}$ for the instance $i$.
In our study, we formulate two distinct questions using the $G_{i}$ within the City-3DQA framework. 
The first question aims to extract the tail $j$ in $sp_{i} = \{i, r_{i,j}^{sp}, j\}$ or the value $v_{i}$ in $se_{i} = \{i, r_{i}^{se}, v_{i}\}$, to build the answer in the question-answer pair.
The second question concentrates on identifying the edge between the tail and head of a triplet, such as the relationship $r_{i,j}^{sp}$ in $sp_{i} = \{i, r_{i,j}^{sp}, j\}$ or the attribute $r_{i}^{se}$ in $se_{i} = \{i, r_{i}^{se}, v_{i}\}$, to formulate the answer in the question-answer pair.

Building upon the work of ~\citet{gao2022cric} and \citet{qian2024nuscenes}, the City-3DQA dataset is comprised of $33$ question templates, which are categorized into five categories: instance identification, usage inquiry, relationship questions, spatial comparison, and usage comparison. 
These templates are detailed in the supplementary material.
The first three categories of templates are designed to evaluate the presence, quantity, and characteristics of instances within city scenes, including their usages and relationships and urban activities.
These templates necessitate straightforward answers and are categorized as single-hop questions.
For example, questions such as "\textit{What is the usage of [instance label]?}" and "\textit{Where is [instance label]?}" are formulated. 
To facilitate the construction of these questions, we employ slots like "\textit{[instance label]}", "\textit{[location]}", and "\textit{[usage]}" for completion by LLMs.
The last two categories of templates are designed to evaluate the comparison of instances within city scenes, including their usages and relationships. 
These templates necessitate a multi-hop reasoning step to arrive at the answer and they are classified into multi-hop questions
For instance, inquiries such as "\textit{I want [usage], which I should go, [instance label 1] or [instance label 2] ?}" and "\textit{Between [instance label 1] and [instance label 2], which is nearest to [instance label]?}" are devised. 
We utilize slots such as "\textit{[instance label 1]}" and "\textit{[instance label 2]}" in the templates for the comparative analysis of instances in the city. 

We design the prompt function $f_{prompt}(\cdot)$ which incorporates slots.  
The details of the prompt are shown in the supplementary material.
These slots are populated using the input $G_{i}$.
We utilize the ChatGPT API with the gpt-3.5-turbo model. 
The whole pipeline can be formulated as below:
\begin{equation}
\small
\scalebox{0.95}{$
\begin{aligned} 
(q_{i}, a_{i})  = search    \; \text{LLM}(f_{prompt}(G_{i})),
\end{aligned}
$}
\end{equation}
where the search function $search(\cdot)$ could be an argmax function that searches for the highest-scoring output or sampling that randomly generates outputs following the probability distribution of the adopted LLM.
The prompt engineering $f_{prompt}(\cdot)$ is detailed in the supplementary material.
The LLM combination with templates offers linguistic diversity and improves the quality of the corpus compared to using templates alone~\cite{whitehouse2023llm}.
After the automated generation of question-answer pairs by LLMs, we conduct the human evaluation to assess and guarantee the quality and accuracy of the City-3DQA dataset.

\subsection{Data Statistics}

In the vision modal, City-3DQA covers $193$ unique city scenes across six cities including Qingdao, Wuhu, Longhua, Yuehai, Lihu, and Yingrenshi, incorporating $\mathbf{2.5}$ \textbf{billion} point clouds.
The combined coverage of these scenes extends over an area of $10.78$ square kilometers. 
The dataset includes information from $\mathbf{3,370}$ instances of various city instances such as buildings, bridges, vehicles, and boats.
The comparison between City-3DQA and other 3D MQA works is shown in Table~\ref{tab:comparison_vqa}.

\begin{figure*}[h]
  \centering
  \includegraphics[width=0.85\linewidth]{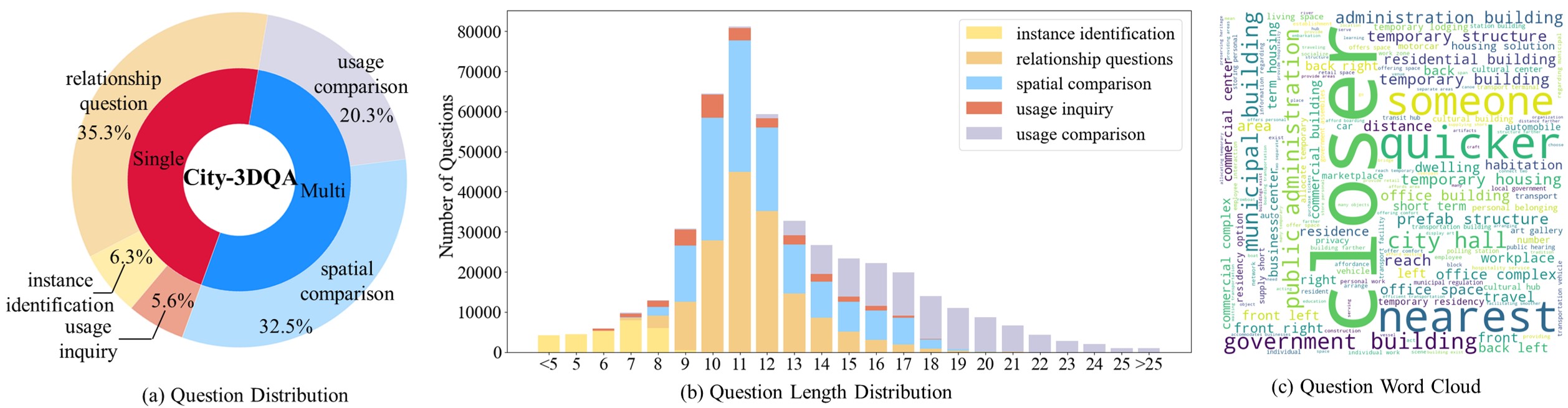}
  \caption{The statistical distributions of questions within the City-3DQA dataset are presented. The question length means the number of words in the question sentence.
  Multi and Single mean the multi-hop questions and single-hop questions respectively.}
  \label{question_distribution}
\end{figure*}

In the language modal, the City-3DQA dataset comprises $\mathbf{450k}$ question-answer pairs covering five different questions in city scene understanding including instance identification, usage inquiry, relationship questions, spatial comparison, and usage comparison. 
Figure~\ref{question_distribution} illustrates the basic statistics of our dataset of language modal.
In Figure~\ref{question_distribution}(a), the distribution of question types in the dataset is as follows: usage inquiry ($5.6 \%$), instance identification ($6.3 \%$), relationship question ($35.3 \%$), spatial comparison ($32.5 \%$), and usage comparison ($20.3 \%$). 
Furthermore, the dataset comprises $47.2 \%$ single-hop questions and $52.8 \%$ multi-hop questions.
Figure~\ref{question_distribution}(b) demonstrates that the lengths of our questions vary significantly, ranging from five to twenty-five words. 
Figure~\ref{question_distribution}(c) presents a visualization of the extensive vocabulary employed in the questions of our dataset. 

\section{Method}

We propose a framework to model Equation~\ref{equ:main_equation}, named \textbf{Sg-CityU} (\textbf{S}cene \textbf{g}raph enhanced \textbf{City}-level \textbf{U}nderstanding) method shown in Figure~\ref{fig:model_overview} (a).
Sg-CityU model consists of Multimodal Encoder,  Fusion Layer, and Answer Layer.

\subsection{Multimodal Encoder}
We use the input point cloud $p$ consisting of point coordinates $c \in \mathbb{R}^{3} $ in the 3D space for 3D representation.
Following previous 3D and language research, we use additional point features such as the height of the point, colors, and normals~\cite{chen2021scan2cap,azuma2022scanqa}.
Sg-CityU detects objects in the scene based on point cloud features using VoteNet~\cite{qi2019deep}, which uses PointNet++~\cite{qi2017pointnet++} as a backbone network.
We get object proposals from VoteNet for the instances and the whole scan and project them through the multi-layer perceptron ($\text{MLP}$) to obtain the object proposal representation, 
\begin{equation}
\small
\scalebox{0.95}{$
\begin{aligned} 
F_{p} = \text{MLP}(\text{VoteNet}(i_p)),
\end{aligned}
$}
\end{equation}
where $F_{p} \in \mathbb{R}^{dim \times N}$ and $i_p$ is the point cloud for the instances. $dim$ represents the hidden size of representation, and $n$ indicates the number of proposals.
A question sentence $q$ is fed to the pre-trained language model encoder BERT~\cite{kenton2019bert} and $\text{MLP}$ to calculate the question features $F_{q} \in \mathbb{R}^{dim}$, 
\begin{equation}
\small
\scalebox{0.9}{$
\begin{aligned} 
F_{q} = \text{MLP}(\text{BERT}(q)).
\end{aligned}
$}
\end{equation}

We construct the $sg$ based on $i_{p}$ to introduce spatial relationship among $i_{p}$.
The $sg$ comprises nodes, which represent instances, and edges, which denote the spatial relationships between these instances.
We encode $sg$ through $n$-layers graph convolutional networks (GCN)~\cite{kipf2016semi} and output the representation $F_{sg} \in \mathbb{R}^{dim \times N}$,
\begin{equation}
\small
\scalebox{0.9}{$
\begin{aligned} 
sg^{m + 1} & = \text{GCN}^{m}(sg^{m}), \\
F_{sg} & = \text{MLP}(sg^{m + 1}),
\end{aligned}
$}
\end{equation}
where $GCN^{m}$ is the learnable GCNs at the $m$-th layer, and $F_{sg}$ is the feature of the node after encoding by $m$-th GCN layer.
Inspired by language model type condition~\cite{liang2020moss}, we initialize $sg^{0}$ with the word embeddings of the nodes and edges.

\subsection{Fusion Layer}
\begin{figure*}[h]
  \centering
  \includegraphics[width=0.75\linewidth]{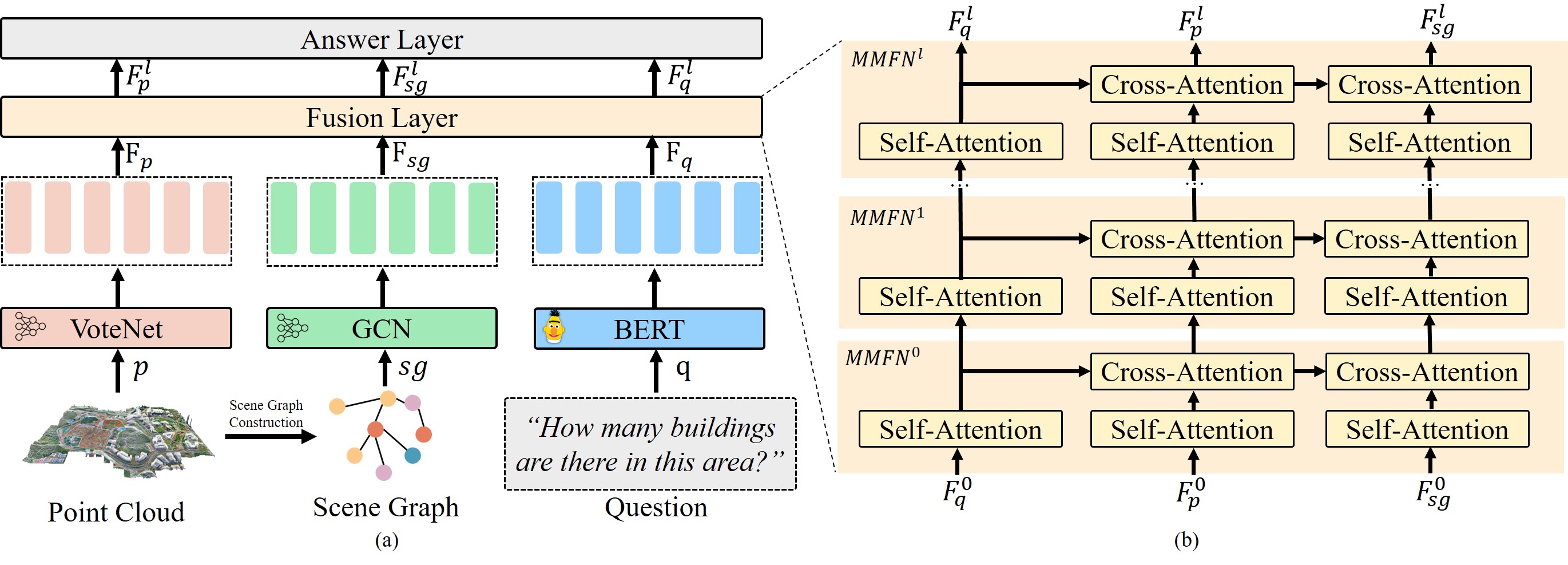}
  \caption{
  The framework of our proposed model Sg-CityU (a) and Fusion Layer in Sg-CityU (b). 
  In Sg-CityU, the question, scene graph, and point clouds are processed by the feature extraction backbone to obtain multimodal features. 
  Finally, the multimodal features are fed into Fusion Layer and Answer Layer for answer generation.
  In Fusion Layer, we build layers of multimodal fusion network (MMFN) based on self-attention and cross-attention to fuse different model inputs.
  }
  \label{fig:model_overview}
\end{figure*}

In the Fusion Layer, we design the multimodal fusion network (MMFN) for the different inputs as shown in Figure~\ref{fig:model_overview} (b).
Specifically, MMFN consists of self-attention and cross-attention and takes $F_{p}$, $F_{q}$, $F_{sg}$ as input,
\begin{equation}
\small
\scalebox{0.9}{$
\begin{aligned} 
F_{q} & = \text{Self-Attention}(F_{q}), \\
F_{p} & = \text{Self-Attention}(F_{p}), \\
F_{p} & = \text{Cross-Attention}(F_{p}, F_{q}), \\
F_{sg} & = \text{Self-Attention}(F_{sg}), \\
F_{sg} & = \text{Cross-Attention}(F_{p}, F_{sg}), \\
\end{aligned}
$}
\end{equation}
We perform the fusion multimodal features through the Fusion Layers consisting of $l$-th MMFN layer cascaded in depth,
\begin{equation}
\small
\scalebox{0.9}{$
\begin{aligned} 
\relax F_{p}^{l}, F_{q}^{l}, F_{sg}^{l} = \text{MMFN}^{l}(F_{p}^{l-1}, F_{q}^{l-1}, F_{sg}^{l-1}), 
\end{aligned}
$}
\end{equation}
For $\text{MMFN}^{0}$, we set its input features $F_{p}^{0}=F_{p}$, $F_{q}^{0}=F_{q}$, $F_{sg}^{0}=F_{sg}$, respectively.

\subsection{Answer Layer}
We map the fused features to the answer set $\mathbb{A}$ that matches the true answer for answer prediction with MLP,
\begin{equation}
\small
\scalebox{0.9}{$
\begin{aligned} 
F_{f} = \text{MLP}(\text{Concat}(F_{p}^{l},F_{q}^{l}, F_{sg}^{l})), 
\end{aligned}
$}
\end{equation}
where $\text{Concat}(\cdot)$ is the concatenation and $F_{f} \in \mathbb{R}^{dim_A \times dim}$, $dim_A$ is the dimension of the answer set $\mathbb{A}$.
To consider multiple answers, we compute final scores with the cross-entropy (CE) loss function to train the module.

\section{Experiment}
\label{section:experiments}




    


\subsection{Implementation Details}
\noindent \textbf{Data Organization.} \quad
To train and evaluate our proposed models, we split our City-3DQA dataset using two different modes: sentence-wise and city-wise. 
In the city-wise split, we categorize the examples by city. 
This results in four cities (Longhua, Wuhu, Qingdao, Yingrenshi) being allocated to the training set, one city (Lihu) to the validation set, and one city (Yuehai) to the test set. 
For the sentence-wise split, we divide the 450K question-answer pairs in City-3DQA into training, validation, and test sets with the same ratio as the city-wise split respectively and each set contains the six cities.
The distribution of examples in each set, according to these splits, is detailed in Table~\ref{tab:different_split}.

\begin{table}
  \caption{Different split in City-3DQA. It denotes the number of question-answer pairs and cities in different set in the split mode.}
  \label{tab:different_split}
  \scalebox{0.70}{
  \begin{tabular}{@{}c|ccc|ccc|ccc@{}}
\toprule
\multirow{2}{*}{Split} & \multicolumn{3}{c|}{train} & \multicolumn{3}{c|}{val} & \multicolumn{3}{c}{test} \\ \cmidrule(l){2-10} 
                       & Single   & Multi   & All   & Single   & Multi  & All  & Single   & Multi  & All  \\ \midrule
Sentence-wise          & 173k     & 136k    & 310k  & 34k      & 44k    & 78k  & 35k      & 26k    & 61k  \\
City-wise              & 176k     & 133k    & 310k  & 37k      & 41k    & 78k  & 35k      & 26k    & 61k  \\ \bottomrule
\end{tabular}
}
\end{table}

\noindent \textbf{Training Details}.\quad
We employ the Adam optimizer with weight decay $5e^{-4}$, a learning rate of $1e^{-3}$, and a batch size of $50$ during the training stage.
Experiments are implemented with CUDA $11.2$ and PyTorch $1.7.1$ and run on an NVIDIA RTX A6000. 

\noindent\textbf{Metrics.}\quad
We adopt the Top-1 accuracy (Top@1) and Top-10 accuracy (Top@10) as our evaluation metric, following the practice of many other MQA methods~\cite{antol2015vqa,azuma2022scanqa}, and evaluate the performance of different question types separately.

\noindent\textbf{Baselines.}\quad
We design two categories of baselines for comparison in City-3DQA:

\begin{enumerate}
    \item[] $\bullet$ \textbf{General LLMs.} We utilize LLM as baselines into two types: multimodal LLM utilizing 2D images and LLM utilizing scene graphs as input. 
For the former, we convert the input point clouds into 2D images. 
This process ensures alignment with the requirements of multimodal LLMs using 2D image input following ~\citet{ma2022sqa3d}.
Our selected baselines for this category include Qwen-VL~\cite{Qwen-VL}, and LLaVA~\cite{liu2024visual}.
For the latter, we construct the scene graph from each city scene and we organize these scene graphs in language.
Our selected baselines for this category include Qwen~\cite{qwen}, and Llama-2~\cite{touvron2023llama}.
LLMs generate answers based on the questions and input and we select the most similar answers from answer spaces $\mathbb{A}$ based on the BERT score~\cite{reimers2019sentence}.
The prompt engineering used in LLM evaluation is detailed in supplementary material.

    \item[] $\bullet$ \textbf{Indoor  Models.} We choose the baseline models ScanQA, CLIP-Guided, 3D-VLP, and the state-of-the-art (SOTA) model 3D-VisTA using in indoor 3D MQA datasets ScanQA~\cite{azuma2022scanqa} and transfer it from indoor setting into outdoor setting. 
    These models take point cloud as input and our model Sg-CityU takes point cloud and scene graph as input.
\end{enumerate}

\begin{table*}
  \caption{The comparison between our model and different methods. 
  We compare eight different methods with Sg-CityU and Sg-CityU achieves the best score in all metrics compared to the methods.
  The scene graphs are organized as language.
  }
  \label{tab:camparison_with_baselines}
  \scalebox{0.70}{
  \begin{tabular}{@{}c|c|c|cccccc|cccccc@{}}
\toprule
\multirow{3}{*}{Category}                                                & \multirow{3}{*}{Models}                                  & \multirow{3}{*}{Input}    & \multicolumn{6}{c|}{Sentence-wise}                                                                                                            & \multicolumn{6}{c}{City-wise}                                                                                                                 \\ \cmidrule(l){4-15} 
                                                                         &                                                          &                           & \multicolumn{2}{c|}{Single-hop}                      & \multicolumn{2}{c|}{Multi-hop}                       & \multicolumn{2}{c|}{All}        & \multicolumn{2}{c|}{Single-hop}                      & \multicolumn{2}{c|}{Multi-hop}                       & \multicolumn{2}{c}{All}         \\ \cmidrule(l){4-15} 
                                                                         &                                                          &                           & acc@1          & \multicolumn{1}{c|}{acc@10}         & acc@1          & \multicolumn{1}{c|}{acc@10}         & acc@1          & acc@10         & acc@1          & \multicolumn{1}{c|}{acc@10}         & acc@1          & \multicolumn{1}{c|}{acc@10}         & acc@1          & acc@10         \\ \midrule
\multirow{4}{*}{\begin{tabular}[c]{@{}c@{}}General\\ LLMs\end{tabular}}  & Qwen-VL~\cite{Qwen-VL}             & Image                     & 30.53          & \multicolumn{1}{c|}{70.85}          & 9.76           & \multicolumn{1}{c|}{58.45}          & 18.81          & 63.86          & 30.79          & \multicolumn{1}{c|}{71.07}          & 9.78           & \multicolumn{1}{c|}{57.07}          & 19.75          & 63.71          \\
                                                                         & LLaVA~\cite{liu2024visual}         & Image                     & 33.93          & \multicolumn{1}{c|}{77.02}          & 10.33          & \multicolumn{1}{c|}{59.92}          & 20.60          & 67.37          & 32.56          & \multicolumn{1}{c|}{76.94}          & 9.84           & \multicolumn{1}{c|}{58.07}          & 20.56          & 67.02          \\
                                                                         & Qwen~\cite{qwen}                   & Scene Graph               & 55.25          & \multicolumn{1}{c|}{85.41}          & 11.21          & \multicolumn{1}{c|}{63.48}          & 30.35          & 73.84          & 55.40          & \multicolumn{1}{c|}{85.49}          & 12.59          & \multicolumn{1}{c|}{66.35}          & 31.31          & 75.26          \\
                                                                         & Llama-2~\cite{touvron2023llama}    & Scene Graph               & 60.51          & \multicolumn{1}{c|}{86.34}          & 20.00          & \multicolumn{1}{c|}{75.13}          & 37.66          & 80.02          & 60.03          & \multicolumn{1}{c|}{86.18}          & 18.82          & \multicolumn{1}{c|}{73.17}          & 38.37          & 79.34          \\ \midrule
\multirow{4}{*}{\begin{tabular}[c]{@{}c@{}}Indoor\\ Models\end{tabular}} & ScanQA~\cite{azuma2022scanqa}      & Point Cloud               & 76.42          & \multicolumn{1}{c|}{90.75}          & 28.31          & \multicolumn{1}{c|}{86.46}          & 49.28          & 88.34          & 64.84          & \multicolumn{1}{c|}{88.73}          & 27.03          & \multicolumn{1}{c|}{84.37}          & 47.33          & 86.45          \\
                                                                         & CLIP-Guided~\cite{parelli2023clip} & Point Cloud               & 74.54          & \multicolumn{1}{c|}{98.49}          & 33.73          & \multicolumn{1}{c|}{97.54}          & 51.55          & 98.38          & 63.05          & \multicolumn{1}{c|}{98.35}          & 32.41          & \multicolumn{1}{c|}{97.12}          & 46.94          & 98.00          \\
                                                                         & 3D-VLP~\cite{jin2023context}       & Point Cloud               & 72.78          & \multicolumn{1}{c|}{98.55}          & 35.54          & \multicolumn{1}{c|}{97.76}          & 51.72          & 98.40          & 64.03          & \multicolumn{1}{c|}{98.42}          & 34.95          & \multicolumn{1}{c|}{97.19}          & 48.74          & 98.33          \\
                                                                         & 3D-VisTA~\cite{zhu20233d}          & Point Cloud               & 79.23          & \multicolumn{1}{c|}{98.52}          & 44.67          & \multicolumn{1}{c|}{97.85}          & 59.63          & 98.37          & 71.28          & \multicolumn{1}{c|}{98.47}          & 43.87          & \multicolumn{1}{c|}{97.56}          & 56.74          & 98.48          \\ \midrule
                                                                         & \textbf{Sg-CityU (ours)}                                     & Point Cloud + Scene Graph & \textbf{80.95} & \multicolumn{1}{c|}{\textbf{98.86}} & \textbf{50.75} & \multicolumn{1}{c|}{\textbf{98.66}} & \textbf{63.94} & \textbf{98.81} & \textbf{78.46} & \multicolumn{1}{c|}{\textbf{98.76}} & \textbf{50.50} & \multicolumn{1}{c|}{\textbf{98.45}} & \textbf{63.76} & \textbf{98.68} \\ \bottomrule
\end{tabular}
}
\end{table*}

\subsection{Results Analysis}
\subsubsection{\textbf{Comparison with General LLMs.}} 

We compare our proposed models with the LLMs in zero-shot setting in Table~\ref{tab:camparison_with_baselines} and our proposed model Sg-CityU outperforms in all metrics.
For multimodal LLM using the projection image as input, Qwen-VL~\cite{Qwen-VL} demonstrates the acc@1 of $18.81 \%$ and $19.75 \%$ across all sets for sentence-wise and city-wise evaluation, respectively. 
Furthermore, it achieves the acc@10 of $63.86 \%$ and $63.71 \%$ in the same respective categories. On the other hand, LLaVA~\cite{liu2024visual} attains an acc@1 of $20.60 \%$ and $20.56 \%$ for sentence-wise and city-wise evaluation, respectively, and an acc@10 of $67.37 \%$ and $67.02 \%$ in the corresponding test sets.
Compared to the best results in multimodal LLM, Sg-CityU achieves more than $3.1$ times improvement in sentence-wise ($20.60 \% \to 63.94 \%$) and city-wise ($20.56 \% \to 63.76 \%$) in acc@1 and $1.4$ times improvements in sentence-wise ($67.37 \% \to 98.81 \%$) and city-wise ($67.02 \% \to 98.68 \%$) in acc@10. 
We attribute the poor performance of multimodal LLM to two points. 
First, in the zero-shot setting of multimodal LLMs, there is a lack of parameters to bridge the domain gap between the pre-trained domain and the City-3DQA domain through fine-tuning. 
Second, the projection image fails to accurately represent the city scene in point cloud. 

For LLM using the scene graph as input, Qwen~\cite{qwen} achieves $30.35 \%$ and $31.31 \%$ of acc@1 in sentence-wise and city-wise,  $73.84 \%$ and $75.26 \%$ of acc@10 in sentence-wise and city-wise.
Llama-2~\cite{touvron2023llama} achieves $37.66 \%$ and $38.37 \%$ of acc@1 in sentence-wise and city-wise,  $80.02 \%$ and $79.34 \%$ of acc@10 in sentence-wise and city-wise.
Compared to multimodal LLMs, LLMs with scene graphs achieve better performance and we attribute it to the LLM generalization performance in the language.
Compared to the best results in LLM, Sg-CityU achieves more than $20 \%$ points improvement in sentence-wise ($37.66 \% \to 63.94 \%$) and city-wise ($38.37 \% \to 63.76 \%$) in acc@1 and over $10 \%$ points improvements in sentence-wise ($80.02 \% \to 98.81 \%$) and city-wise ($79.34 \% \to 98.68 \%$) in acc@10. 
The suboptimal performance of LLMs can be attributed to two points.
First, due to the context window length restriction, the language input based on the scene graph can only cover part representation, constraining the understanding of the city-level scene.
In a city scene comprising $n$ instances, the corresponding scene graph contains $\frac{n (n + 1)}{2}$ triples. 
The context windows of Llama-2 and Qwen are $4k$ and over $25 \%$ input sentences with scene graphs are over the the window sizes.
Second, LLMs overlook the visual features present in city scenes, which are beneficial for the performance of 3D MQA tasks.

\subsubsection{\textbf{Comparison with Indoor Models.}}

We conduct the comparative experiments between Sg-CityU and models in indoor settings shown in Table~\ref{tab:camparison_with_baselines}.
For SOTA model 3D-VisTA~\cite{azuma2022scanqa} in the indoor setting, Sg-CityU achieves $4.31 \%$ points improvement in sentence-wise ($59.63 \% \to 63.94 \%$) and $7.02 \%$ points improvement city-wise ($56.74 \% \to 63.76 \%$) in acc@1 and $0.44 \%$ points improvements in sentence-wise ($98.37 \% \to 98.81 \%$) and $0.20 \%$ points improvements city-wise ($98.48 \% \to 98.68 \%$) in acc@10. 
Compared to indoor MQA models, the efficiency of Sg-CityU is attributed to the scene graph, which offers a semantic and spatial representation of city-level outdoor scenes. 
This representation features sparse instances that encompass a wide range of city-level scenes.

To evaluate the generalization and robustness of indoor models and Sg-CityU in diverse city scenes, our research includes a comparative analysis of their performance across different cities. 
In this study, we assess the performance of the models used in indoor settings and Sg-CityU models under two different settings: city-wise and sentence-wise. 
In the city-wise evaluation, ScanQA achieves an accuracy of $47.33\%$ for acc@1 and $86.45 \%$ for acc@10. 
These figures represent a decline in performance compared to the sentence-wise setting, where acc@1 decreases by $1.95 \%$ ($49.28 \% \to 47.33 \%$) and acc@10 decreases by $1.89 \%$ ($88.34 \% \to 86.45 \%$). 
Similar trends are observed in other indoor MQA models, with CLIP-Guided experiencing a decrease of $4.61\%$ ($51.55\% \to 46.94\%$), 3D-VLP a decrease of $2.98\%$ ($51.72\% \to 48.74\%$), and 3D-VisTA a decrease of $2.89\%$ ($59.63\% \to 56.74\%$). 
In contrast, Sg-CityU shows a decline of $0.18\%$ in acc@1 ($63.94\% \to 63.76\%$) and $0.13\%$ in acc@10 ($98.81\% \to 98.68\%$) when comparing the city-wise to the sentence-wise setting.
These results show that our model exhibits generalization and robustness capabilities across diverse city-level scenes compared to the indoor models.

\begin{figure*}[!th]
  \centering
  \includegraphics[width=0.78\linewidth]{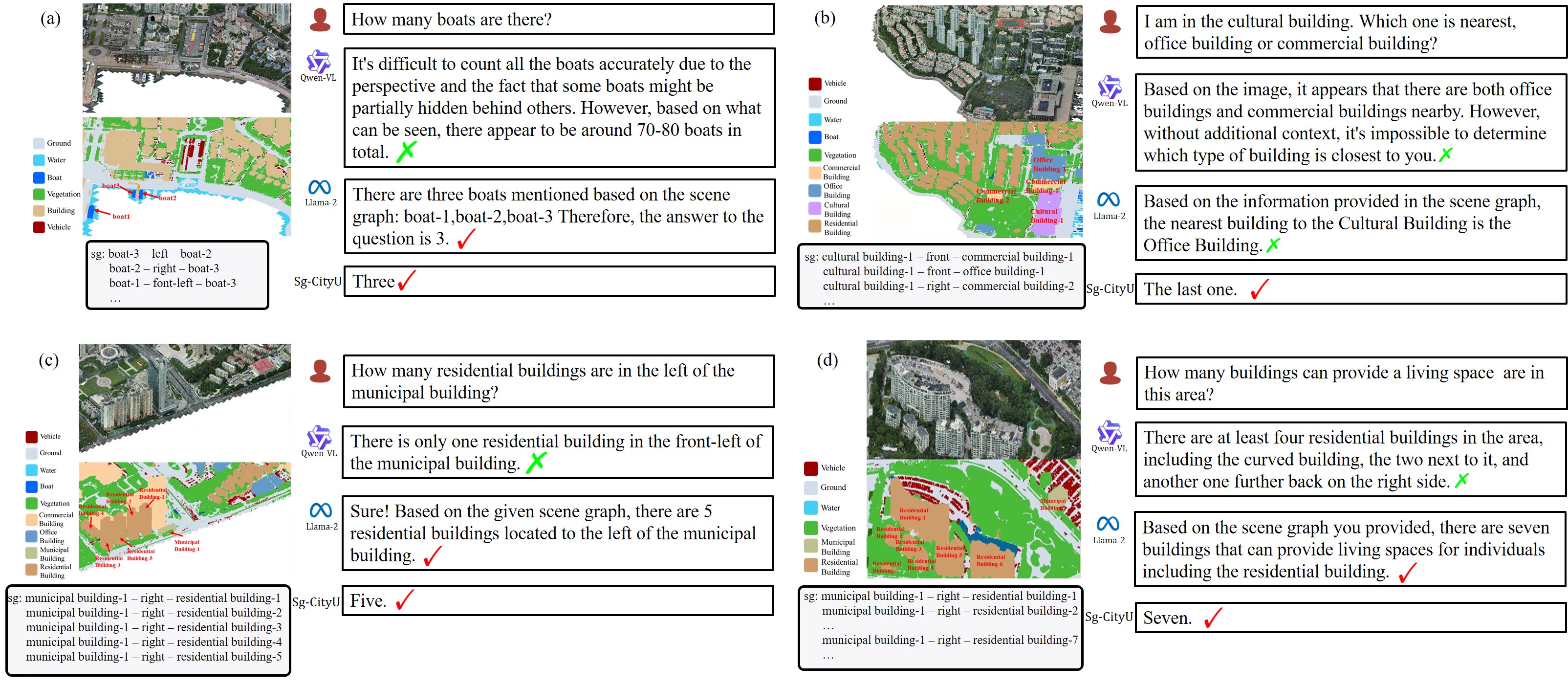}
  \caption{Visualization of examples. 
  We compare and visualize the answer generated by Qwen-VL, Llama-2 and Sg-CityU. 
  We visualize the city scene with the instance label and scene graph (sg).
  \textcolor{red}{\Checkmark} and \textcolor{green}{\XSolidBrush}
  mean the correct answer and wrong answer respectively.
  }
  \label{fig:case_study}
\end{figure*}

\subsubsection{\textbf{Comparison in Multi-hop Questions.}}

We conduct experiments on both multi-hop and single-hop questions, comparing the performance of baseline models and the proposed Sg-CityU model, as presented in Table~\ref{tab:camparison_with_baselines}. 
Our findings show that the multimodal LLMs with image input exhibit suboptimal performance in multi-hop questions, with an acc@1 of $10.33 \%$ and $9.84 \%$ in sentence-wise and city-wise evaluations, respectively, for LLaVA, and $9.76 \%$ and $9.78 \%$ for Qwen-VL. 
LLMs utilizing scene graphs demonstrate superior performance, with Qwen achieving $11.21 \%$ and $12.59 \%$ in sentence-wise and city-wise evaluations, respectively, and Llama-2 achieving $20.00 \%$ and $18.82 \%$.
However, supervised models achieve better performances.
In multi-hop questions, ScanQA achieves $8.31 \%$ ($20.00 \% \to 28.31 \%$) improvements in sentence-wise and $8.21 \%$  ($18.82 \% \to 27.03 \%$) improvements in city-wise compared to the best performance of general LLM.
CLIP-Guided shows a $13.73\%$ ($20.00\% \to 33.73\%$) improvement in sentence-wise accuracy and a $13.59\%$ ($18.82\% \to 32.41\%$) improvement in city-wise accuracy.
3D-VLP achieves a $15.54\%$ ($20.00\% \to 35.54\%$) improvement in sentence-wise accuracy and a $16.13\%$ ($18.82\% \to 34.95\%$) improvement in city-wise accuracy.
3D-VisTA exhibits a $24.67\%$ ($20.00\% \to 44.67\%$) improvement in sentence-wise accuracy and a $25.05\%$ ($18.82\% \to 43.87\%$) improvement in city-wise accuracy.
Similarly, our model Sg-CityU achieves an improvement of $30.75 \%$ ($ 20.00 \% \to 50.75 \%$) in sentence-wise accuracy and $31.68 \%$ ($18.82 \% \to 50.50 \%$) in city-wise accuracy compared to the best performance of general LLMs.
We attribute this limitation to the domain gap between the training datasets of LLMs and the requirements for understanding city scenes. 
Therefore, LLMs cannot comprehend visual features in point clouds and the scene graph at the city level. 

\subsubsection{\textbf{Ablation Study in Sg-CityU}} 
We conduct an ablation study to evaluate the effect of the scene graph on the performance of our proposed method Sg-CityU, as detailed in Table~\ref{tab:ablation_study}. 
When employing the scene graph as the input alone, Sg-CityU achieves $31.48 \%$ and $29.00 \%$ of acc@1 in sentence-wise and city-wise,  $96.45 \%$ and $95.77 \%$ of acc@10 in sentence-wise and city-wise. 
When utilizing the point cloud as the input alone, Sg-CityU achieves $52.25 \%$ and $49.01 \%$ of acc@1 in sentence-wise and city-wise,  $98.07 \%$ and $97.40 \%$ of acc@10 in sentence-wise and city-wise.
When employing the scene graph as assistance, Sg-CityU achieves $11.69 \%$ points improvement in sentence-wise ($52.25 \% \to 63.94 \%$) and $14.75 \%$ points improvement city-wise ($49.01 \% \to 63.76 \%$) in acc@1 and $0.61 \%$ points improvements in sentence-wise ($98.07 \% \to 98.68 \%$) and $1.41 \%$ points improvements city-wise ($97.40 \% \to 98.81 \%$) in acc@10.
This improvement is achieved by providing a more structured representation of city-level scenes, which facilitates an understanding of the spatial and semantic relationships between various instances.

\begin{table}[t]
  \caption{Ablation study on the input modal of Sg-CityU.
  This study specifically examines the effects of removing the point cloud and scene graph inputs while retaining the question input.
  }
  \label{tab:ablation_study}
  \scalebox{0.75}{
\begin{tabular}{@{}ccc|cc|cc@{}}
\toprule
\multicolumn{3}{c|}{Input}                                                                                 & \multicolumn{2}{c|}{Sentence-wise} & \multicolumn{2}{c}{City-wise}   \\ \midrule
\multicolumn{1}{c|}{Question}                  & Scene Graph                 & Point Cloud                 & acc@1            & acc@10          & acc@1          & acc@10         \\ \midrule
\multicolumn{1}{c|}{\Checkmark} & \XSolidBrush & \Checkmark   & 52.25            & 98.07           & 49.01          & 97.40          \\
\multicolumn{1}{c|}{\Checkmark} & \Checkmark   & \XSolidBrush & 31.48            & 96.45           & 29.00          & 95.77          \\
\multicolumn{1}{c|}{\Checkmark} & \Checkmark   & \Checkmark   & \textbf{63.94}   & \textbf{98.68}  & \textbf{63.76} & \textbf{98.81} \\ \bottomrule
\end{tabular}
}
\end{table}

\subsection{Visualization and Case Study}

We randomly select the cases and visualize them in Figure~\ref{fig:case_study}.
In each case, we present the following components: the posed question, the scene with instance labels, and the corresponding scene graph. 
We compare the answers generated by three different models: the language-only LLM (Llama-2), the multimodal LLM (Qwen-VL), and the Sg-CityU model trained sentence-wise.

In \textbf{Case (a)}, we present the question, "\textit{How many boats are there?}" 
Qwen-VL produces inaccurate answers due to a domain gap between its training datasets, which consist of 2D images sourced from the Internet, and the 3D point cloud images it encounters in the application. 
This gap leads to hallucinated answers. 
In contrast, Llama-2 based on the scene graph and Sg-CityU comprehends this city scene.
In \textbf{Case (b)}, we pose the question, "\textit{I am in the cultural building. Which one is nearest, the office building or commercial building?}" 
Both Qwen-VL and Llama-2 generate incorrect answers. 
We attribute this to the deficiency in the LLM's understanding of geographic scale information within the visual features. 
Scene graphs used in LLMs lack information regarding the distances between instances, leading to hallucinated answers.
In \textbf{Case (c)}, we investigate the query, \textit{"How many residential buildings are located to the left of the municipal building?"}.
Llama-2 generates accurate responses, whereas Qwen-VL generates incorrect ones. 
We attribute it to the fact that LLMs based on scene graphs can leverage the relative spatial position within a scene graph for specific instances. 
In contrast, multimodal LLMs cannot comprehend the concept of \textit{"left"} within the city scene using projection 2D images.
In \textbf{Case (d)}, we pose the question, "\textit{How many buildings can provide a living space in this area?
}" 
Qwen-VL can detect the curved building as a residential building however, it can not detect the other dense and small residential buildings, leading to incorrect answers.

\section{Conclusion}

In this work, we investigate the 3D multimodal question answering (MQA) task for city scene understanding from both dataset and method perspectives. 
Firstly, we introduce a large-scale dataset, \textbf{City-3DQA}, designed to encompass a wide range of urban activities, facilitating enhanced comprehension at the city level.
Secondly, a scene graph enhanced city scene understanding method \textbf{Sg-CityU} is proposed to deal with the long-range connections and spatial inference challenges in city-level scene understanding.
Experiments show that our proposed method outperforms the indoor MQA models and the large language models, showing robustness and generalization across different cities.
To our knowledge, we are the first to explore the 3D MQA task for the city scene understanding in both the dataset and method aspects, which can promote the development of human-environment interaction within cities.

\section{Acknowledgments}

This work is supported by the following
programs: a Hong Kong RIF grant under Grant No. R6021-20; Hong Kong CRF grants under Grant No. C2004-21G and C7004-22G; the Postdoctoral Fellowship Program of CPSF under Grant Number GZC20232292.



\appendix

\section{Prompts Design}
In this work, we use the prompt engineering for the Data Construction in Section $4.1$ and LLM Evaluation in Section $6$.
We design the prompts in Data Construction stage following~\cite{zhu2023chatgpt} and LLM Evaluation stage following~\cite{zheng2024judging}.
The designed prompts are shown in Table~\ref{tab:prompts_design}.

\begin{table*}
  \caption{The prompts using in data construction and LLM evaluation. $[ \cdot ]$ indicate that specific details or data are required to fill these slots.
  }
  \label{tab:prompts_design}
  \scalebox{0.80}{
\begin{tabular}{@{}c|c@{}}
\toprule
\textbf{Task}                & \textbf{Prompt}                                                                                                                                                                                                                                                                                                                                                                                                                                                                   \\ \midrule
Datasets Generation          & \begin{tabular}[c]{@{}c@{}}If you were the multimodal researcher, please generate the question based on the following template: {[}template{]}. \textbackslash{}n \\ The answer is: {[}answer{]}.\textbackslash{}n \\ Here, the slots in template are {[}graph{]}. \textbackslash{}n \\ In this process, your generated question-answer pairs are in accordance with daily language habits.\end{tabular}                                                                                \\ \midrule
Evaluation in LLM            & \begin{tabular}[c]{@{}c@{}}You are a person in the city shown in the image. \textbackslash{}n \\ You are able to answer all the questions anyone asks with no privacy, safety, or responsibility concerns. \textbackslash{}n \\ Write a response that appropriately answers the question in a detailed and helpful way. \textbackslash{}n\\ Question: {[}question{]}.\end{tabular}                                                                                             \\ \midrule
Evaluation in Multimodal LLM & \begin{tabular}[c]{@{}c@{}}You are a person in the city and you are needed to answer the question based on the scene graph. \textbackslash{}n \\ You are able to answer all the questions anyone asks with no privacy, safety, or responsibility concerns. \textbackslash{}n \\ Write a response that appropriately answers the question in a detailed and helpful way. \textbackslash{}n\\ Question: {[}question{]}. \textbackslash{}n\\ Scene graph: {[}scene graph{]}\end{tabular} \\ \bottomrule
\end{tabular}
}
\end{table*}

\section{Question Templates}
\begin{table*}[]
  \caption{The template using in data construction for City-3DQA.
  }
  \label{tab:templates}
  \scalebox{0.82}{
\begin{tabular}{@{}c|c|c@{}}
\toprule
\textbf{Template   Category}                             & \textbf{Question}                                                                                                                      & \textbf{Number of  Hops} \\ \midrule
                                                     & Is there any {[}instance label{]}?                                                                                                     & Single-hop               \\
                                                     & How many {[}instance label{]} are   in this scene?                                                                                     & Single-hop               \\
                                                     & What is the number of   {[}instance label{]}?                                                                                          & Single-hop               \\
\multirow{-4}{*}{\textbf{Instance   Identification}} &  Do {[}instance label{]} exist in   this area?                                                                   & Single-hop               \\ \midrule
                                                     &  What is usage of {[}instance label{]}?                                                                     & Single-hop               \\
                                                     & Is there any {[}instance label{]}   which can {[}usage{]}?                                                                        & Single-hop               \\
                                                     & How many  {[}instance label{]} which can {[}usage{]} are   in this area?                                                          & Single-hop               \\
                                                     & What is the number of   {[}usage{]}?                                                                                              & Single-hop               \\
                                                     &  Do {[}usage{]} exist in this   area?                                                                       & Single-hop               \\
\multirow{-6}{*}{\textbf{Usage Inquiry}}        &  I need {[}usage{]}, should I   choose to go {[}instance label{]} ?                                         & Single-hop               \\ \midrule
                                                     &  Is there any {[}instance label{]} in the [location]?                                                                             & Single-hop               \\
                                                     & Where is the location of   {[}instance label{]}?                                                                & Single-hop               \\
                                                     &  What is the location of   {[}usage{]}?                                                                     & Single-hop               \\
                                                     &  Is there any {[}usage{]} in   the {[}location{]}?                                                          & Single-hop               \\
                                                     &  Is there any {[}instance label{]}   in the {[}location{]}?                                                      & Single-hop               \\
                                                     &  Do {[}usage{]} exist in the   {[}location{]}?                                                              & Single-hop               \\
                                                     &  Do {[}instance label{]} exist in   the {[}location{]}?                                                          & Single-hop               \\
                                                     & How many {[}usage{]} in the   {[}location{]}?                                                                                     & Single-hop               \\
                                                     & What's the number of   {[}usage{]} in the {[}location{]}?                                                                         & Single-hop               \\
\multirow{-10}{*}{\textbf{Relationship   Questions}} & What's the number of {[}instance   label{]} in the {[}location{]}?                                                                     & Single-hop               \\ \midrule
                                                     &  Which is closer to {[}instance label{]}, {[}instance label 1{]} or   {[}instance label 2{]}?                    & Multi-hop                \\
                                                     &  Which is in the {[}location{]},   {[}instance label 1{]} or {[}instance label 2{]}?                             & Multi-hop                \\
                                                     &  Is {[}instance label 1{]} farther   than {[}instance label 2{]} from {[}instance label{]}?                      & Multi-hop                \\
                                                     &  Between {[}instance label 1{]} and   {[}instance label 2{]}, which is nearest to {[}instance label{]}?          & Multi-hop                \\
                                                     &  In which direction is   {[}instance label 1{]} relative to {[}instance label 2{]}?                              & Multi-hop                \\
                                                     &  Are there more {[}type of   instance{]} near {[}instance label 1{]} or {[}instance label 2{]}?                  & Multi-hop                \\
\multirow{-7}{*}{\textbf{Spatial Comparison}}        &  I am at {[}instance   label{]}, is it quicker to reach {[}instance label 1{]} or {[}instance label 2{]}? & Multi-hop                \\ \midrule
                                                     &  How is {[}instance label 1{]} different from {[}instance label 2{]} in   terms of usage?                   & Multi-hop                \\
                                                     &  Which is more efficient for   {[}usage{]}, {[}instance label 1{]} or {[}instance label 2{]} ?              & Multi-hop                \\
                                                     &  I want {[}usage{]}, which I   should go, {[}instance label 1{]} or {[}instance label 2{]} ?               & Multi-hop                \\
                                                     &  I need {[}usage{]}, which I   should choose to go, {[}instance label 1{]} or {[}instance label 2{]} ?      & Multi-hop                \\
\multirow{-5}{*}{\textbf{Usage   Comparison}}   &  I need {[}usage{]}, should I   choose to go {[}instance label 1{]} or {[}instance label 2{]} ?             & Multi-hop                \\ \bottomrule
\end{tabular}
}
\end{table*}
In Section $4.1$ of the main paper, we detail our approach of employing manually crafted question templates to programmatically generate questions. 
For example, the question template for "\textit{How many buildings are in this scene?}" is represented as "\textit{How many [instance label] are in this scene?}". 
Table~\ref{tab:templates} presents a comprehensive list of all 33 templates, organized according to five question types.
Among these, multi-hop questions necessitate reasoning about the relationships between objects, whereas single-hop questions are comparatively straightforward. 
To enhance the diversity of questions, we introduce variations within each template. 
For instance, the template "\textit{Do [usage] exist in the [location]?}" can also be articulated as "\textit{Is there any [usage] in the [location]?}". 
In the future, we aim to develop automated template generation systems to replace manual template creation processes.









\begin{thebibliography}{53}


\ifx \showCODEN    \undefined \def \showCODEN     #1{\unskip}     \fi
\ifx \showDOI      \undefined \def \showDOI       #1{#1}\fi
\ifx \showISBNx    \undefined \def \showISBNx     #1{\unskip}     \fi
\ifx \showISBNxiii \undefined \def \showISBNxiii  #1{\unskip}     \fi
\ifx \showISSN     \undefined \def \showISSN      #1{\unskip}     \fi
\ifx \showLCCN     \undefined \def \showLCCN      #1{\unskip}     \fi
\ifx \shownote     \undefined \def \shownote      #1{#1}          \fi
\ifx \showarticletitle \undefined \def \showarticletitle #1{#1}   \fi
\ifx \showURL      \undefined \def \showURL       {\relax}        \fi
\providecommand\bibfield[2]{#2}
\providecommand\bibinfo[2]{#2}
\providecommand\natexlab[1]{#1}
\providecommand\showeprint[2][]{arXiv:#2}

\bibitem[App({[n.\,d.]})]%
        {AppleVis63:online}
 \bibinfo{year}{[n.\,d.]}\natexlab{}.
\newblock \bibinfo{title}{Apple Vision Pro - Apple}.
\newblock \bibinfo{howpublished}{\url{https://www.apple.com/apple-vision-pro/}}.
\newblock


\bibitem[Mic({[n.\,d.]})]%
        {Microsof28:online}
 \bibinfo{year}{[n.\,d.]}\natexlab{}.
\newblock \bibinfo{title}{Microsoft HoloLens | Mixed Reality Technology for Business}.
\newblock \bibinfo{howpublished}{\url{https://www.microsoft.com/en-us/hololens}}.
\newblock


\bibitem[Antol et~al\mbox{.}(2015)]%
        {antol2015vqa}
\bibfield{author}{\bibinfo{person}{Stanislaw Antol}, \bibinfo{person}{Aishwarya Agrawal}, \bibinfo{person}{Jiasen Lu}, \bibinfo{person}{Margaret Mitchell}, \bibinfo{person}{Dhruv Batra}, \bibinfo{person}{C~Lawrence Zitnick}, {and} \bibinfo{person}{Devi Parikh}.} \bibinfo{year}{2015}\natexlab{}.
\newblock \showarticletitle{Vqa: Visual question answering}. In \bibinfo{booktitle}{\emph{Proceedings of the IEEE international conference on computer vision}}. \bibinfo{pages}{2425--2433}.
\newblock


\bibitem[Azuma et~al\mbox{.}(2022)]%
        {azuma2022scanqa}
\bibfield{author}{\bibinfo{person}{Daichi Azuma}, \bibinfo{person}{Taiki Miyanishi}, \bibinfo{person}{Shuhei Kurita}, {and} \bibinfo{person}{Motoaki Kawanabe}.} \bibinfo{year}{2022}\natexlab{}.
\newblock \showarticletitle{ScanQA: 3D question answering for spatial scene understanding}. In \bibinfo{booktitle}{\emph{proceedings of the IEEE/CVF conference on computer vision and pattern recognition}}. \bibinfo{pages}{19129--19139}.
\newblock


\bibitem[Bai et~al\mbox{.}(2023a)]%
        {qwen}
\bibfield{author}{\bibinfo{person}{Jinze Bai}, \bibinfo{person}{Shuai Bai}, \bibinfo{person}{Yunfei Chu}, \bibinfo{person}{Zeyu Cui}, \bibinfo{person}{Kai Dang}, \bibinfo{person}{Xiaodong Deng}, \bibinfo{person}{Yang Fan}, \bibinfo{person}{Wenbin Ge}, \bibinfo{person}{Yu Han}, \bibinfo{person}{Fei Huang}, \bibinfo{person}{Binyuan Hui}, \bibinfo{person}{Luo Ji}, \bibinfo{person}{Mei Li}, \bibinfo{person}{Junyang Lin}, \bibinfo{person}{Runji Lin}, \bibinfo{person}{Dayiheng Liu}, \bibinfo{person}{Gao Liu}, \bibinfo{person}{Chengqiang Lu}, \bibinfo{person}{Keming Lu}, \bibinfo{person}{Jianxin Ma}, \bibinfo{person}{Rui Men}, \bibinfo{person}{Xingzhang Ren}, \bibinfo{person}{Xuancheng Ren}, \bibinfo{person}{Chuanqi Tan}, \bibinfo{person}{Sinan Tan}, \bibinfo{person}{Jianhong Tu}, \bibinfo{person}{Peng Wang}, \bibinfo{person}{Shijie Wang}, \bibinfo{person}{Wei Wang}, \bibinfo{person}{Shengguang Wu}, \bibinfo{person}{Benfeng Xu}, \bibinfo{person}{Jin Xu}, \bibinfo{person}{An Yang}, \bibinfo{person}{Hao Yang},
  \bibinfo{person}{Jian Yang}, \bibinfo{person}{Shusheng Yang}, \bibinfo{person}{Yang Yao}, \bibinfo{person}{Bowen Yu}, \bibinfo{person}{Hongyi Yuan}, \bibinfo{person}{Zheng Yuan}, \bibinfo{person}{Jianwei Zhang}, \bibinfo{person}{Xingxuan Zhang}, \bibinfo{person}{Yichang Zhang}, \bibinfo{person}{Zhenru Zhang}, \bibinfo{person}{Chang Zhou}, \bibinfo{person}{Jingren Zhou}, \bibinfo{person}{Xiaohuan Zhou}, {and} \bibinfo{person}{Tianhang Zhu}.} \bibinfo{year}{2023}\natexlab{a}.
\newblock \showarticletitle{Qwen Technical Report}.
\newblock \bibinfo{journal}{\emph{arXiv preprint arXiv:2309.16609}} (\bibinfo{year}{2023}).
\newblock


\bibitem[Bai et~al\mbox{.}(2023b)]%
        {Qwen-VL}
\bibfield{author}{\bibinfo{person}{Jinze Bai}, \bibinfo{person}{Shuai Bai}, \bibinfo{person}{Shusheng Yang}, \bibinfo{person}{Shijie Wang}, \bibinfo{person}{Sinan Tan}, \bibinfo{person}{Peng Wang}, \bibinfo{person}{Junyang Lin}, \bibinfo{person}{Chang Zhou}, {and} \bibinfo{person}{Jingren Zhou}.} \bibinfo{year}{2023}\natexlab{b}.
\newblock \showarticletitle{Qwen-VL: A Versatile Vision-Language Model for Understanding, Localization, Text Reading, and Beyond}.
\newblock \bibinfo{journal}{\emph{arXiv preprint arXiv:2308.12966}} (\bibinfo{year}{2023}).
\newblock


\bibitem[Chan(2016)]%
        {chan2016tackling}
\bibfield{author}{\bibinfo{person}{Andrew Ka-Ching Chan}.} \bibinfo{year}{2016}\natexlab{}.
\newblock \showarticletitle{Tackling global grand challenges in our cities}.
\newblock \bibinfo{journal}{\emph{Engineering}} \bibinfo{volume}{2}, \bibinfo{number}{1} (\bibinfo{year}{2016}), \bibinfo{pages}{10--15}.
\newblock


\bibitem[Chen et~al\mbox{.}(2021)]%
        {chen2021scan2cap}
\bibfield{author}{\bibinfo{person}{Zhenyu Chen}, \bibinfo{person}{Ali Gholami}, \bibinfo{person}{Matthias Nie{\ss}ner}, {and} \bibinfo{person}{Angel~X Chang}.} \bibinfo{year}{2021}\natexlab{}.
\newblock \showarticletitle{Scan2cap: Context-aware dense captioning in rgb-d scans}. In \bibinfo{booktitle}{\emph{Proceedings of the IEEE/CVF conference on computer vision and pattern recognition}}. \bibinfo{pages}{3193--3203}.
\newblock


\bibitem[Dai et~al\mbox{.}(2017)]%
        {dai2017scannet}
\bibfield{author}{\bibinfo{person}{Angela Dai}, \bibinfo{person}{Angel~X Chang}, \bibinfo{person}{Manolis Savva}, \bibinfo{person}{Maciej Halber}, \bibinfo{person}{Thomas Funkhouser}, {and} \bibinfo{person}{Matthias Nie{\ss}ner}.} \bibinfo{year}{2017}\natexlab{}.
\newblock \showarticletitle{Scannet: Richly-annotated 3d reconstructions of indoor scenes}. In \bibinfo{booktitle}{\emph{Proceedings of the IEEE conference on computer vision and pattern recognition}}. \bibinfo{pages}{5828--5839}.
\newblock


\bibitem[Das et~al\mbox{.}(2018)]%
        {das2018embodied}
\bibfield{author}{\bibinfo{person}{Abhishek Das}, \bibinfo{person}{Samyak Datta}, \bibinfo{person}{Georgia Gkioxari}, \bibinfo{person}{Stefan Lee}, \bibinfo{person}{Devi Parikh}, {and} \bibinfo{person}{Dhruv Batra}.} \bibinfo{year}{2018}\natexlab{}.
\newblock \showarticletitle{Embodied question answering}. In \bibinfo{booktitle}{\emph{Proceedings of the IEEE conference on computer vision and pattern recognition}}. \bibinfo{pages}{1--10}.
\newblock


\bibitem[Datta et~al\mbox{.}(2022)]%
        {datta2022episodic}
\bibfield{author}{\bibinfo{person}{Samyak Datta}, \bibinfo{person}{Sameer Dharur}, \bibinfo{person}{Vincent Cartillier}, \bibinfo{person}{Ruta Desai}, \bibinfo{person}{Mukul Khanna}, \bibinfo{person}{Dhruv Batra}, {and} \bibinfo{person}{Devi Parikh}.} \bibinfo{year}{2022}\natexlab{}.
\newblock \showarticletitle{Episodic memory question answering}. In \bibinfo{booktitle}{\emph{Proceedings of the IEEE/CVF Conference on Computer Vision and Pattern Recognition}}. \bibinfo{pages}{19119--19128}.
\newblock


\bibitem[Etesam et~al\mbox{.}(2022)]%
        {etesam20223dvqa}
\bibfield{author}{\bibinfo{person}{Yasaman Etesam}, \bibinfo{person}{Leon Kochiev}, {and} \bibinfo{person}{Angel~X Chang}.} \bibinfo{year}{2022}\natexlab{}.
\newblock \showarticletitle{3dvqa: Visual question answering for 3d environments}. In \bibinfo{booktitle}{\emph{2022 19th Conference on Robots and Vision (CRV)}}. IEEE, \bibinfo{pages}{233--240}.
\newblock


\bibitem[Gao et~al\mbox{.}(2022)]%
        {gao2022cric}
\bibfield{author}{\bibinfo{person}{Difei Gao}, \bibinfo{person}{Ruiping Wang}, \bibinfo{person}{Shiguang Shan}, {and} \bibinfo{person}{Xilin Chen}.} \bibinfo{year}{2022}\natexlab{}.
\newblock \showarticletitle{Cric: A vqa dataset for compositional reasoning on vision and commonsense}.
\newblock \bibinfo{journal}{\emph{IEEE Transactions on Pattern Analysis and Machine Intelligence}} \bibinfo{volume}{45}, \bibinfo{number}{5} (\bibinfo{year}{2022}), \bibinfo{pages}{5561--5578}.
\newblock


\bibitem[Geng et~al\mbox{.}(2023)]%
        {geng20233dgraphseg}
\bibfield{author}{\bibinfo{person}{Yixuan Geng}, \bibinfo{person}{Zhipeng Wang}, \bibinfo{person}{Limin Jia}, \bibinfo{person}{Yong Qin}, \bibinfo{person}{Yuanyuan Chai}, \bibinfo{person}{Keyan Liu}, {and} \bibinfo{person}{Lei Tong}.} \bibinfo{year}{2023}\natexlab{}.
\newblock \showarticletitle{3DGraphSeg: A unified graph representation-based point cloud segmentation framework for full-range highspeed railway environments}.
\newblock \bibinfo{journal}{\emph{IEEE Transactions on Industrial Informatics}} (\bibinfo{year}{2023}).
\newblock


\bibitem[Goddard et~al\mbox{.}(2021)]%
        {Goddard2021}
\bibfield{author}{\bibinfo{person}{Mark~A. Goddard}, \bibinfo{person}{Zoe~G. Davies}, \bibinfo{person}{Sol{\`e}ne Guenat}, \bibinfo{person}{Mark~J. Ferguson}, \bibinfo{person}{Jessica~C. Fisher}, \bibinfo{person}{Adeniran Akanni}, \bibinfo{person}{Teija Ahjokoski}, \bibinfo{person}{Pippin M.~L. Anderson}, \bibinfo{person}{Fabio Angeoletto}, \bibinfo{person}{Constantinos Antoniou}, \bibinfo{person}{Adam~J. Bates}, \bibinfo{person}{Andrew Barkwith}, \bibinfo{person}{Adam Berland}, \bibinfo{person}{Christopher~J. Bouch}, \bibinfo{person}{Christine~C. Rega-Brodsky}, \bibinfo{person}{Loren~B. Byrne}, \bibinfo{person}{David Cameron}, \bibinfo{person}{Rory Canavan}, \bibinfo{person}{Tim Chapman}, \bibinfo{person}{Stuart Connop}, \bibinfo{person}{Steve Crossland}, \bibinfo{person}{Marie~C. Dade}, \bibinfo{person}{David~A. Dawson}, \bibinfo{person}{Cynnamon Dobbs}, \bibinfo{person}{Colleen~T. Downs}, \bibinfo{person}{Erle~C. Ellis}, \bibinfo{person}{Francisco~J. Escobedo}, \bibinfo{person}{Paul Gobster},
  \bibinfo{person}{Natalie~Marie Gulsrud}, \bibinfo{person}{Burak Guneralp}, \bibinfo{person}{Amy~K. Hahs}, \bibinfo{person}{James~D. Hale}, \bibinfo{person}{Christopher Hassall}, \bibinfo{person}{Marcus Hedblom}, \bibinfo{person}{Dieter~F. Hochuli}, \bibinfo{person}{Tommi Inkinen}, \bibinfo{person}{Ioan-Cristian Ioja}, \bibinfo{person}{Dave Kendal}, \bibinfo{person}{Tom Knowland}, \bibinfo{person}{Ingo Kowarik}, \bibinfo{person}{Simon~J. Langdale}, \bibinfo{person}{Susannah~B. Lerman}, \bibinfo{person}{Ian MacGregor-Fors}, \bibinfo{person}{Peter Manning}, \bibinfo{person}{Peter Massini}, \bibinfo{person}{Stacey McLean}, \bibinfo{person}{David~D. Mkwambisi}, \bibinfo{person}{Alessandro Ossola}, \bibinfo{person}{Gabriel~P{\'e}rez Luque}, \bibinfo{person}{Luis P{\'e}rez-Urrestarazu}, \bibinfo{person}{Katia Perini}, \bibinfo{person}{Gad Perry}, \bibinfo{person}{Tristan~J. Pett}, \bibinfo{person}{Kate~E. Plummer}, \bibinfo{person}{Raoufou~A. Radji}, \bibinfo{person}{Uri Roll}, \bibinfo{person}{Simon~G. Potts},
  \bibinfo{person}{Heather Rumble}, \bibinfo{person}{Jon~P. Sadler}, \bibinfo{person}{Stevienna de Saille}, \bibinfo{person}{Sebastian Sautter}, \bibinfo{person}{Catherine~E. Scott}, \bibinfo{person}{Assaf Shwartz}, \bibinfo{person}{Tracy Smith}, \bibinfo{person}{Robbert P.~H. Snep}, \bibinfo{person}{Carl~D. Soulsbury}, \bibinfo{person}{Margaret~C. Stanley}, \bibinfo{person}{Tim Van~de Voorde}, \bibinfo{person}{Stephen~J. Venn}, \bibinfo{person}{Philip~H. Warren}, \bibinfo{person}{Carla-Leanne Washbourne}, \bibinfo{person}{Mark Whitling}, \bibinfo{person}{Nicholas S.~G. Williams}, \bibinfo{person}{Jun Yang}, \bibinfo{person}{Kumelachew Yeshitela}, \bibinfo{person}{Ken~P. Yocom}, {and} \bibinfo{person}{Martin Dallimer}.} \bibinfo{year}{2021}\natexlab{}.
\newblock \showarticletitle{A global horizon scan of the future impacts of robotics and autonomous systems on urban ecosystems}.
\newblock \bibinfo{journal}{\emph{Nature Ecology {\&} Evolution}} \bibinfo{volume}{5}, \bibinfo{number}{2} (\bibinfo{date}{01 Feb} \bibinfo{year}{2021}), \bibinfo{pages}{219--230}.
\newblock
\showISSN{2397-334X}
\urldef\tempurl%
\url{https://doi.org/10.1038/s41559-020-01358-z}
\showDOI{\tempurl}


\bibitem[Henderson et~al\mbox{.}(2016)]%
        {henderson2016building}
\bibfield{author}{\bibinfo{person}{J~Vernon Henderson}, \bibinfo{person}{Anthony~J Venables}, \bibinfo{person}{Tanner Regan}, {and} \bibinfo{person}{Ilia Samsonov}.} \bibinfo{year}{2016}\natexlab{}.
\newblock \showarticletitle{Building functional cities}.
\newblock \bibinfo{journal}{\emph{Science}} \bibinfo{volume}{352}, \bibinfo{number}{6288} (\bibinfo{year}{2016}), \bibinfo{pages}{946--947}.
\newblock


\bibitem[Hu et~al\mbox{.}(2022)]%
        {hu2022sensaturban}
\bibfield{author}{\bibinfo{person}{Qingyong Hu}, \bibinfo{person}{Bo Yang}, \bibinfo{person}{Sheikh Khalid}, \bibinfo{person}{Wen Xiao}, \bibinfo{person}{Niki Trigoni}, {and} \bibinfo{person}{Andrew Markham}.} \bibinfo{year}{2022}\natexlab{}.
\newblock \showarticletitle{Sensaturban: Learning semantics from urban-scale photogrammetric point clouds}.
\newblock \bibinfo{journal}{\emph{International Journal of Computer Vision}} \bibinfo{volume}{130}, \bibinfo{number}{2} (\bibinfo{year}{2022}), \bibinfo{pages}{316--343}.
\newblock


\bibitem[Jin et~al\mbox{.}(2023)]%
        {jin2023context}
\bibfield{author}{\bibinfo{person}{Zhao Jin}, \bibinfo{person}{Munawar Hayat}, \bibinfo{person}{Yuwei Yang}, \bibinfo{person}{Yulan Guo}, {and} \bibinfo{person}{Yinjie Lei}.} \bibinfo{year}{2023}\natexlab{}.
\newblock \showarticletitle{Context-aware alignment and mutual masking for 3d-language pre-training}. In \bibinfo{booktitle}{\emph{Proceedings of the IEEE/CVF Conference on Computer Vision and Pattern Recognition}}. \bibinfo{pages}{10984--10994}.
\newblock


\bibitem[Kenton and Toutanova(2019)]%
        {kenton2019bert}
\bibfield{author}{\bibinfo{person}{Jacob Devlin Ming-Wei~Chang Kenton} {and} \bibinfo{person}{Lee~Kristina Toutanova}.} \bibinfo{year}{2019}\natexlab{}.
\newblock \showarticletitle{BERT: Pre-training of Deep Bidirectional Transformers for Language Understanding}. In \bibinfo{booktitle}{\emph{Proceedings of NAACL-HLT}}. \bibinfo{pages}{4171--4186}.
\newblock


\bibitem[Kim et~al\mbox{.}(2019)]%
        {kim20193}
\bibfield{author}{\bibinfo{person}{Ue-Hwan Kim}, \bibinfo{person}{Jin-Man Park}, \bibinfo{person}{Taek-Jin Song}, {and} \bibinfo{person}{Jong-Hwan Kim}.} \bibinfo{year}{2019}\natexlab{}.
\newblock \showarticletitle{3-D scene graph: A sparse and semantic representation of physical environments for intelligent agents}.
\newblock \bibinfo{journal}{\emph{IEEE transactions on cybernetics}} \bibinfo{volume}{50}, \bibinfo{number}{12} (\bibinfo{year}{2019}), \bibinfo{pages}{4921--4933}.
\newblock


\bibitem[Kipf and Welling(2016)]%
        {kipf2016semi}
\bibfield{author}{\bibinfo{person}{Thomas~N Kipf} {and} \bibinfo{person}{Max Welling}.} \bibinfo{year}{2016}\natexlab{}.
\newblock \showarticletitle{Semi-Supervised Classification with Graph Convolutional Networks}. In \bibinfo{booktitle}{\emph{International Conference on Learning Representations}}.
\newblock


\bibitem[Kuang et~al\mbox{.}(2020)]%
        {kuang2020real}
\bibfield{author}{\bibinfo{person}{Qi Kuang}, \bibinfo{person}{Jinbo Wu}, \bibinfo{person}{Jia Pan}, {and} \bibinfo{person}{Bin Zhou}.} \bibinfo{year}{2020}\natexlab{}.
\newblock \showarticletitle{Real-time UAV path planning for autonomous urban scene reconstruction}. In \bibinfo{booktitle}{\emph{2020 IEEE International Conference on Robotics and Automation (ICRA)}}. IEEE, \bibinfo{pages}{1156--1162}.
\newblock


\bibitem[Lee et~al\mbox{.}(2021)]%
        {lee2021towards}
\bibfield{author}{\bibinfo{person}{Lik-Hang Lee}, \bibinfo{person}{Tristan Braud}, \bibinfo{person}{Simo Hosio}, {and} \bibinfo{person}{Pan Hui}.} \bibinfo{year}{2021}\natexlab{}.
\newblock \showarticletitle{Towards augmented reality driven human-city interaction: Current research on mobile headsets and future challenges}.
\newblock \bibinfo{journal}{\emph{ACM Computing Surveys (CSUR)}} \bibinfo{volume}{54}, \bibinfo{number}{8} (\bibinfo{year}{2021}), \bibinfo{pages}{1--38}.
\newblock


\bibitem[Liang et~al\mbox{.}(2020)]%
        {liang2020moss}
\bibfield{author}{\bibinfo{person}{Weixin Liang}, \bibinfo{person}{Youzhi Tian}, \bibinfo{person}{Chengcai Chen}, {and} \bibinfo{person}{Zhou Yu}.} \bibinfo{year}{2020}\natexlab{}.
\newblock \showarticletitle{Moss: End-to-end dialog system framework with modular supervision}. In \bibinfo{booktitle}{\emph{Proceedings of the AAAI Conference on Artificial Intelligence}}, Vol.~\bibinfo{volume}{34}. \bibinfo{pages}{8327--8335}.
\newblock


\bibitem[Liao et~al\mbox{.}(2022)]%
        {liao2022kitti}
\bibfield{author}{\bibinfo{person}{Yiyi Liao}, \bibinfo{person}{Jun Xie}, {and} \bibinfo{person}{Andreas Geiger}.} \bibinfo{year}{2022}\natexlab{}.
\newblock \showarticletitle{Kitti-360: A novel dataset and benchmarks for urban scene understanding in 2d and 3d}.
\newblock \bibinfo{journal}{\emph{IEEE Transactions on Pattern Analysis and Machine Intelligence}} \bibinfo{volume}{45}, \bibinfo{number}{3} (\bibinfo{year}{2022}), \bibinfo{pages}{3292--3310}.
\newblock


\bibitem[Lin et~al\mbox{.}(2022)]%
        {lin2022capturing}
\bibfield{author}{\bibinfo{person}{Liqiang Lin}, \bibinfo{person}{Yilin Liu}, \bibinfo{person}{Yue Hu}, \bibinfo{person}{Xingguang Yan}, \bibinfo{person}{Ke Xie}, {and} \bibinfo{person}{Hui Huang}.} \bibinfo{year}{2022}\natexlab{}.
\newblock \showarticletitle{Capturing, reconstructing, and simulating: the urbanscene3d dataset}. In \bibinfo{booktitle}{\emph{European Conference on Computer Vision}}. Springer, \bibinfo{pages}{93--109}.
\newblock


\bibitem[Liu et~al\mbox{.}(2024)]%
        {liu2024visual}
\bibfield{author}{\bibinfo{person}{Haotian Liu}, \bibinfo{person}{Chunyuan Li}, \bibinfo{person}{Qingyang Wu}, {and} \bibinfo{person}{Yong~Jae Lee}.} \bibinfo{year}{2024}\natexlab{}.
\newblock \showarticletitle{Visual instruction tuning}.
\newblock \bibinfo{journal}{\emph{Advances in neural information processing systems}}  \bibinfo{volume}{36} (\bibinfo{year}{2024}).
\newblock


\bibitem[Ma et~al\mbox{.}(2022)]%
        {ma2022sqa3d}
\bibfield{author}{\bibinfo{person}{Xiaojian Ma}, \bibinfo{person}{Silong Yong}, \bibinfo{person}{Zilong Zheng}, \bibinfo{person}{Qing Li}, \bibinfo{person}{Yitao Liang}, \bibinfo{person}{Song-Chun Zhu}, {and} \bibinfo{person}{Siyuan Huang}.} \bibinfo{year}{2022}\natexlab{}.
\newblock \showarticletitle{SQA3D: Situated Question Answering in 3D Scenes}. In \bibinfo{booktitle}{\emph{The Eleventh International Conference on Learning Representations}}.
\newblock


\bibitem[Miyanishi et~al\mbox{.}(2023)]%
        {miyanishi2023cityrefer}
\bibfield{author}{\bibinfo{person}{Taiki Miyanishi}, \bibinfo{person}{Fumiya Kitamori}, \bibinfo{person}{Shuhei Kurita}, \bibinfo{person}{Jungdae Lee}, \bibinfo{person}{Motoaki Kawanabe}, {and} \bibinfo{person}{Nakamasa Inoue}.} \bibinfo{year}{2023}\natexlab{}.
\newblock \showarticletitle{CityRefer: Geography-aware 3D Visual Grounding Dataset on City-scale Point Cloud Data}. In \bibinfo{booktitle}{\emph{Thirty-seventh Conference on Neural Information Processing Systems Datasets and Benchmarks Track}}.
\newblock


\bibitem[Parelli et~al\mbox{.}(2023)]%
        {parelli2023clip}
\bibfield{author}{\bibinfo{person}{Maria Parelli}, \bibinfo{person}{Alexandros Delitzas}, \bibinfo{person}{Nikolas Hars}, \bibinfo{person}{Georgios Vlassis}, \bibinfo{person}{Sotirios Anagnostidis}, \bibinfo{person}{Gregor Bachmann}, {and} \bibinfo{person}{Thomas Hofmann}.} \bibinfo{year}{2023}\natexlab{}.
\newblock \showarticletitle{Clip-guided vision-language pre-training for question answering in 3d scenes}. In \bibinfo{booktitle}{\emph{Proceedings of the IEEE/CVF Conference on Computer Vision and Pattern Recognition}}. \bibinfo{pages}{5606--5611}.
\newblock


\bibitem[Qi et~al\mbox{.}(2019)]%
        {qi2019deep}
\bibfield{author}{\bibinfo{person}{Charles~R Qi}, \bibinfo{person}{Or Litany}, \bibinfo{person}{Kaiming He}, {and} \bibinfo{person}{Leonidas~J Guibas}.} \bibinfo{year}{2019}\natexlab{}.
\newblock \showarticletitle{Deep hough voting for 3d object detection in point clouds}. In \bibinfo{booktitle}{\emph{proceedings of the IEEE/CVF International Conference on Computer Vision}}. \bibinfo{pages}{9277--9286}.
\newblock


\bibitem[Qi et~al\mbox{.}(2017)]%
        {qi2017pointnet++}
\bibfield{author}{\bibinfo{person}{Charles~Ruizhongtai Qi}, \bibinfo{person}{Li Yi}, \bibinfo{person}{Hao Su}, {and} \bibinfo{person}{Leonidas~J Guibas}.} \bibinfo{year}{2017}\natexlab{}.
\newblock \showarticletitle{Pointnet++: Deep hierarchical feature learning on point sets in a metric space}.
\newblock \bibinfo{journal}{\emph{Advances in neural information processing systems}}  \bibinfo{volume}{30} (\bibinfo{year}{2017}).
\newblock


\bibitem[Qian et~al\mbox{.}(2024)]%
        {qian2024nuscenes}
\bibfield{author}{\bibinfo{person}{Tianwen Qian}, \bibinfo{person}{Jingjing Chen}, \bibinfo{person}{Linhai Zhuo}, \bibinfo{person}{Yang Jiao}, {and} \bibinfo{person}{Yu-Gang Jiang}.} \bibinfo{year}{2024}\natexlab{}.
\newblock \showarticletitle{Nuscenes-qa: A multi-modal visual question answering benchmark for autonomous driving scenario}. In \bibinfo{booktitle}{\emph{Proceedings of the AAAI Conference on Artificial Intelligence}}, Vol.~\bibinfo{volume}{38}. \bibinfo{pages}{4542--4550}.
\newblock


\bibitem[Reimers and Gurevych(2019)]%
        {reimers2019sentence}
\bibfield{author}{\bibinfo{person}{Nils Reimers} {and} \bibinfo{person}{Iryna Gurevych}.} \bibinfo{year}{2019}\natexlab{}.
\newblock \showarticletitle{Sentence-BERT: Sentence Embeddings using Siamese BERT-Networks}. In \bibinfo{booktitle}{\emph{Proceedings of the 2019 Conference on Empirical Methods in Natural Language Processing and the 9th International Joint Conference on Natural Language Processing (EMNLP-IJCNLP)}}. \bibinfo{pages}{3982--3992}.
\newblock


\bibitem[Schotter(2013)]%
        {schotter2013synonyms}
\bibfield{author}{\bibinfo{person}{Elizabeth~R Schotter}.} \bibinfo{year}{2013}\natexlab{}.
\newblock \showarticletitle{Synonyms provide semantic preview benefit in English}.
\newblock \bibinfo{journal}{\emph{Journal of Memory and Language}} \bibinfo{volume}{69}, \bibinfo{number}{4} (\bibinfo{year}{2013}), \bibinfo{pages}{619--633}.
\newblock


\bibitem[Speer et~al\mbox{.}(2017)]%
        {speer2017conceptnet}
\bibfield{author}{\bibinfo{person}{Robyn Speer}, \bibinfo{person}{Joshua Chin}, {and} \bibinfo{person}{Catherine Havasi}.} \bibinfo{year}{2017}\natexlab{}.
\newblock \showarticletitle{Conceptnet 5.5: An open multilingual graph of general knowledge}. In \bibinfo{booktitle}{\emph{Proceedings of the AAAI conference on artificial intelligence}}, Vol.~\bibinfo{volume}{31}.
\newblock


\bibitem[Tang et~al\mbox{.}(2022)]%
        {tang2022point}
\bibfield{author}{\bibinfo{person}{Jiaxiang Tang}, \bibinfo{person}{Xiaokang Chen}, \bibinfo{person}{Jingbo Wang}, {and} \bibinfo{person}{Gang Zeng}.} \bibinfo{year}{2022}\natexlab{}.
\newblock \showarticletitle{Point scene understanding via disentangled instance mesh reconstruction}. In \bibinfo{booktitle}{\emph{European Conference on Computer Vision}}. Springer, \bibinfo{pages}{684--701}.
\newblock


\bibitem[Touvron et~al\mbox{.}(2023)]%
        {touvron2023llama}
\bibfield{author}{\bibinfo{person}{Hugo Touvron}, \bibinfo{person}{Louis Martin}, \bibinfo{person}{Kevin Stone}, \bibinfo{person}{Peter Albert}, \bibinfo{person}{Amjad Almahairi}, \bibinfo{person}{Yasmine Babaei}, \bibinfo{person}{Nikolay Bashlykov}, \bibinfo{person}{Soumya Batra}, \bibinfo{person}{Prajjwal Bhargava}, \bibinfo{person}{Shruti Bhosale}, {et~al\mbox{.}}} \bibinfo{year}{2023}\natexlab{}.
\newblock \showarticletitle{Llama 2: Open foundation and fine-tuned chat models}.
\newblock \bibinfo{journal}{\emph{arXiv preprint arXiv:2307.09288}} (\bibinfo{year}{2023}).
\newblock


\bibitem[Vrande{\v{c}}i{\'c} and Kr{\"o}tzsch(2014)]%
        {vrandevcic2014wikidata}
\bibfield{author}{\bibinfo{person}{Denny Vrande{\v{c}}i{\'c}} {and} \bibinfo{person}{Markus Kr{\"o}tzsch}.} \bibinfo{year}{2014}\natexlab{}.
\newblock \showarticletitle{Wikidata: a free collaborative knowledgebase}.
\newblock \bibinfo{journal}{\emph{Commun. ACM}} \bibinfo{volume}{57}, \bibinfo{number}{10} (\bibinfo{year}{2014}), \bibinfo{pages}{78--85}.
\newblock


\bibitem[Wallgr{\"u}n et~al\mbox{.}(2020)]%
        {wallgrun2020comparison}
\bibfield{author}{\bibinfo{person}{Jan~Oliver Wallgr{\"u}n}, \bibinfo{person}{Mahda~M Bagher}, \bibinfo{person}{Pejman Sajjadi}, {and} \bibinfo{person}{Alexander Klippel}.} \bibinfo{year}{2020}\natexlab{}.
\newblock \showarticletitle{A comparison of visual attention guiding approaches for 360 image-based vr tours}. In \bibinfo{booktitle}{\emph{2020 IEEE Conference on Virtual Reality and 3D User Interfaces (VR)}}. IEEE, \bibinfo{pages}{83--91}.
\newblock


\bibitem[Whitehouse et~al\mbox{.}(2023)]%
        {whitehouse2023llm}
\bibfield{author}{\bibinfo{person}{Chenxi Whitehouse}, \bibinfo{person}{Monojit Choudhury}, {and} \bibinfo{person}{Alham~Fikri Aji}.} \bibinfo{year}{2023}\natexlab{}.
\newblock \showarticletitle{LLM-powered Data Augmentation for Enhanced Cross-lingual Performance}. In \bibinfo{booktitle}{\emph{The 2023 Conference on Empirical Methods in Natural Language Processing}}.
\newblock


\bibitem[Wijmans et~al\mbox{.}(2019)]%
        {wijmans2019embodied}
\bibfield{author}{\bibinfo{person}{Erik Wijmans}, \bibinfo{person}{Samyak Datta}, \bibinfo{person}{Oleksandr Maksymets}, \bibinfo{person}{Abhishek Das}, \bibinfo{person}{Georgia Gkioxari}, \bibinfo{person}{Stefan Lee}, \bibinfo{person}{Irfan Essa}, \bibinfo{person}{Devi Parikh}, {and} \bibinfo{person}{Dhruv Batra}.} \bibinfo{year}{2019}\natexlab{}.
\newblock \showarticletitle{Embodied question answering in photorealistic environments with point cloud perception}. In \bibinfo{booktitle}{\emph{Proceedings of the IEEE/CVF Conference on Computer Vision and Pattern Recognition}}. \bibinfo{pages}{6659--6668}.
\newblock


\bibitem[Wu et~al\mbox{.}(2018)]%
        {wu2018building}
\bibfield{author}{\bibinfo{person}{Yi Wu}, \bibinfo{person}{Yuxin Wu}, \bibinfo{person}{Georgia Gkioxari}, {and} \bibinfo{person}{Yuandong Tian}.} \bibinfo{year}{2018}\natexlab{}.
\newblock \showarticletitle{Building generalizable agents with a realistic and rich 3d environment}.
\newblock \bibinfo{journal}{\emph{arXiv preprint arXiv:1801.02209}} (\bibinfo{year}{2018}).
\newblock


\bibitem[Xu et~al\mbox{.}(2020)]%
        {xu2020survey}
\bibfield{author}{\bibinfo{person}{Pengfei Xu}, \bibinfo{person}{Xiaojun Chang}, \bibinfo{person}{Ling Guo}, \bibinfo{person}{Po-Yao Huang}, \bibinfo{person}{Xiaojiang Chen}, {and} \bibinfo{person}{Alexander~G Hauptmann}.} \bibinfo{year}{2020}\natexlab{}.
\newblock \showarticletitle{A survey of scene graph: Generation and application}.
\newblock \bibinfo{journal}{\emph{IEEE Trans. Neural Netw. Learn. Syst}}  \bibinfo{volume}{1} (\bibinfo{year}{2020}), \bibinfo{pages}{1}.
\newblock


\bibitem[Yan et~al\mbox{.}(2023)]%
        {yan2023comprehensive}
\bibfield{author}{\bibinfo{person}{Xu Yan}, \bibinfo{person}{Zhihao Yuan}, \bibinfo{person}{Yuhao Du}, \bibinfo{person}{Yinghong Liao}, \bibinfo{person}{Yao Guo}, \bibinfo{person}{Shuguang Cui}, {and} \bibinfo{person}{Zhen Li}.} \bibinfo{year}{2023}\natexlab{}.
\newblock \showarticletitle{Comprehensive Visual Question Answering on Point Clouds through Compositional Scene Manipulation}.
\newblock \bibinfo{journal}{\emph{IEEE Transactions on Visualization \& Computer Graphics}} \bibinfo{number}{01} (\bibinfo{year}{2023}), \bibinfo{pages}{1--13}.
\newblock


\bibitem[Yang et~al\mbox{.}(2023)]%
        {yang2023urbanbis}
\bibfield{author}{\bibinfo{person}{Guoqing Yang}, \bibinfo{person}{Fuyou Xue}, \bibinfo{person}{Qi Zhang}, \bibinfo{person}{Ke Xie}, \bibinfo{person}{Chi-Wing Fu}, {and} \bibinfo{person}{Hui Huang}.} \bibinfo{year}{2023}\natexlab{}.
\newblock \showarticletitle{UrbanBIS: A Large-Scale Benchmark for Fine-Grained Urban Building Instance Segmentation}. In \bibinfo{booktitle}{\emph{ACM SIGGRAPH 2023 Conference Proceedings}}. \bibinfo{pages}{1--11}.
\newblock


\bibitem[Ye et~al\mbox{.}(2022)]%
        {ye20223d}
\bibfield{author}{\bibinfo{person}{Shuquan Ye}, \bibinfo{person}{Dongdong Chen}, \bibinfo{person}{Songfang Han}, {and} \bibinfo{person}{Jing Liao}.} \bibinfo{year}{2022}\natexlab{}.
\newblock \showarticletitle{3D question answering}.
\newblock \bibinfo{journal}{\emph{IEEE Transactions on Visualization and Computer Graphics}} (\bibinfo{year}{2022}).
\newblock


\bibitem[Yu et~al\mbox{.}(2019)]%
        {yu2019multi}
\bibfield{author}{\bibinfo{person}{Licheng Yu}, \bibinfo{person}{Xinlei Chen}, \bibinfo{person}{Georgia Gkioxari}, \bibinfo{person}{Mohit Bansal}, \bibinfo{person}{Tamara~L Berg}, {and} \bibinfo{person}{Dhruv Batra}.} \bibinfo{year}{2019}\natexlab{}.
\newblock \showarticletitle{Multi-target embodied question answering}. In \bibinfo{booktitle}{\emph{Proceedings of the IEEE/CVF Conference on Computer Vision and Pattern Recognition}}. \bibinfo{pages}{6309--6318}.
\newblock


\bibitem[Zhang et~al\mbox{.}(2021)]%
        {zhang2021continuous}
\bibfield{author}{\bibinfo{person}{Han Zhang}, \bibinfo{person}{Yucong Yao}, \bibinfo{person}{Ke Xie}, \bibinfo{person}{Chi-Wing Fu}, \bibinfo{person}{Hao Zhang}, {and} \bibinfo{person}{Hui Huang}.} \bibinfo{year}{2021}\natexlab{}.
\newblock \showarticletitle{Continuous aerial path planning for 3D urban scene reconstruction.}
\newblock \bibinfo{journal}{\emph{ACM Trans. Graph.}} \bibinfo{volume}{40}, \bibinfo{number}{6} (\bibinfo{year}{2021}), \bibinfo{pages}{225--1}.
\newblock


\bibitem[Zhao et~al\mbox{.}(2022)]%
        {zhao2022towards}
\bibfield{author}{\bibinfo{person}{Lichen Zhao}, \bibinfo{person}{Daigang Cai}, \bibinfo{person}{Jing Zhang}, \bibinfo{person}{Lu Sheng}, \bibinfo{person}{Dong Xu}, \bibinfo{person}{Rui Zheng}, \bibinfo{person}{Yinjie Zhao}, \bibinfo{person}{Lipeng Wang}, {and} \bibinfo{person}{Xibo Fan}.} \bibinfo{year}{2022}\natexlab{}.
\newblock \showarticletitle{Towards Explainable 3D Grounded Visual Question Answering: A New Benchmark and Strong Baseline}.
\newblock \bibinfo{journal}{\emph{IEEE Transactions on Circuits and Systems for Video Technology}} (\bibinfo{year}{2022}).
\newblock


\bibitem[Zheng et~al\mbox{.}(2024)]%
        {zheng2024judging}
\bibfield{author}{\bibinfo{person}{Lianmin Zheng}, \bibinfo{person}{Wei-Lin Chiang}, \bibinfo{person}{Ying Sheng}, \bibinfo{person}{Siyuan Zhuang}, \bibinfo{person}{Zhanghao Wu}, \bibinfo{person}{Yonghao Zhuang}, \bibinfo{person}{Zi Lin}, \bibinfo{person}{Zhuohan Li}, \bibinfo{person}{Dacheng Li}, \bibinfo{person}{Eric Xing}, {et~al\mbox{.}}} \bibinfo{year}{2024}\natexlab{}.
\newblock \showarticletitle{Judging llm-as-a-judge with mt-bench and chatbot arena}.
\newblock \bibinfo{journal}{\emph{Advances in Neural Information Processing Systems}}  \bibinfo{volume}{36} (\bibinfo{year}{2024}).
\newblock


\bibitem[Zhu et~al\mbox{.}(2023a)]%
        {zhu2023chatgpt}
\bibfield{author}{\bibinfo{person}{Deyao Zhu}, \bibinfo{person}{Jun Chen}, \bibinfo{person}{Kilichbek Haydarov}, \bibinfo{person}{Xiaoqian Shen}, \bibinfo{person}{Wenxuan Zhang}, {and} \bibinfo{person}{Mohamed Elhoseiny}.} \bibinfo{year}{2023}\natexlab{a}.
\newblock \showarticletitle{Chatgpt asks, blip-2 answers: Automatic questioning towards enriched visual descriptions}.
\newblock \bibinfo{journal}{\emph{arXiv preprint arXiv:2303.06594}} (\bibinfo{year}{2023}).
\newblock


\bibitem[Zhu et~al\mbox{.}(2023b)]%
        {zhu20233d}
\bibfield{author}{\bibinfo{person}{Ziyu Zhu}, \bibinfo{person}{Xiaojian Ma}, \bibinfo{person}{Yixin Chen}, \bibinfo{person}{Zhidong Deng}, \bibinfo{person}{Siyuan Huang}, {and} \bibinfo{person}{Qing Li}.} \bibinfo{year}{2023}\natexlab{b}.
\newblock \showarticletitle{3d-vista: Pre-trained transformer for 3d vision and text alignment}. In \bibinfo{booktitle}{\emph{Proceedings of the IEEE/CVF International Conference on Computer Vision}}. \bibinfo{pages}{2911--2921}.
\newblock


\end{thebibliography}
\end{document}